%% file: HAL_article.tex
\documentclass{article}

\input{preamble-standalone}
\input{preamble-local}

\usepackage{fix2col}

\begin{document}

\makeatletter
\def\url@ukostyle{%
  \@ifundefined{selectfont}{\def\UrlFont{\sf}}{\def\UrlFont{\sf}}}
\makeatother
\urlstyle{uko}

\input{gen/moffet_template}

{\centering
  {\bfseries\Large Compressive Pattern Matching on Multispectral Data\bigskip}

  S. Rousseau\textsuperscript{1},
  D. Helbert\textsuperscript{1},
  P. Carr\'e\textsuperscript{1},
  J. Blanc-Talon\textsuperscript{2}\\

  {\itshape
    \textsuperscript{1}XLIM Laboratory, UMR CNRS 7252, University of Poitiers, France\\
    \textsuperscript{2}DGA/DS/QIS, Bagneux, France\\
  }
}

\newif\ifCLASSOPTIONonecolumn


\begin{abstract}
  We introduce a new constrained minimization problem that performs
  template and pattern detection on a multispectral image in a
  compressive sensing context. We use an original minimization problem
  from Guo and Osher that uses $L_1$ minimization techniques to
  perform template detection in a multispectral image. We first adapt
  this minimization problem to work with compressive sensing data.
  Then we extend it to perform pattern detection using a formal
  transform called the \textit{spectralization} along a pattern. That
  extension brings out the problem of measurement reconstruction. We
  introduce shifted measurements that allow us to reconstruct all the
  measurement with a small overhead and we give an optimality
  constraint for simple patterns. We present numerical results showing
  the performances of the original minimization problem and the
  compressed ones with different measurement rates and applied on
  remotely sensed data.
\end{abstract}


\section{Introduction}
\label{sec:introduction}

The compressive sensing is a recent field of signal processing. It has
been introduced by Donoho~\cite{donoho2006compressed} and Cand\`es,
Romberg and Tao~\cite{candes2006stable}. The main idea is that the
acquisition and the compression steps are performed simultaneously by
taking a limited number of linear measurements on the signal. These
linear measurements are modelled as inner products against a signal
$x$ of size $n$\label{rev:1}:
\begin{equation}
  y_i=\langle\phi_i, x\rangle,\quad i=1,\dots,m
\end{equation}
where $m$ is the number of linear measurements. This can be written
$y=\bPhi x$ where the $\phi_i$'s are the columns of $\bPhi^T$ and
$\bPhi$ is referred to as the sensing matrix. Recovering the signal
$x$ for the linear measurements $y$ is a linear inverse problem that
has more unknowns than equations because $m\leqslant n$. But if the
signal $x$ is sparse in a known basis, it can be recovered from the
measurements. The signal $x$ is sparse in the orthonormal bases
$\bPsi$ if we can decompose $x$ as $\bPsi u$ where $u$ is sparse. In
that case, the following minimization problem involving a $L_0$
norm\footnote{The $L_0$ norm just counts all the nonzero entries.}
recovers $u$.
\begin{equation}\label{eq:13}
  \argmin_u\|u\|_0\quad\textrm{s.t.}\quad\bPhi\bPsi u=y.
\end{equation}
Unfortunately, due to its combinatorial nature, that minimization
problem is intractable. Several methods have been proposed to find an
approximate solution. Greedy approaches have been explored giving
\emph{matching pursuit}-like algorithms (see~\cite{mallat1993matching,
  pati1993orthogonal, Tropp07signalrecovery, blumensath2008gradient,
  davis1997adaptive, donoho2006stable}). Another way to efficiently
solve this problem is to consider the closest convex problem that
involves a $L_1$ norm called the \emph{basis pursuit}
problem~\cite{Chen98atomicdecomposition}
\begin{equation}\label{eq:4}
  \argmin_u\|u\|_1 \quad\textrm{s.t.}\quad\bPhi\bPsi u=y.
\end{equation}
The $L_1$ norm just replaces the $L_0$ norm. It is proved under
certain circumstances that the solutions of both problems coincides.

Multispectral imagery requires the acquisition of each band of the
multispectral image. In this regard, the compressive sensing paradigm
becomes very interesting since the standard acquisition process
generates a huge flow of data and thus requires a costly compression
step. Some physical implementations have already been proposed: the
single-pixel hyperspectral camera~\cite{duarte2008single} based on
digital micromirror device (DMD) or the Coded Aperture Snapshot
Spectral Imaging (CASSI)~\cite{gehm2007single} based on two
dispersive elements. Another benefit of such a new acquisition
paradigm is that the signal does not need to be fully reconstructed
before performing some processing task
(see~\cite{davenport2006detection,wang2008subspace}). That bypassing
technique has been applied to various hyperspectral image processing
such as spectral unmixing ~\cite{li2011compressive, zare2012directly,
  golbabaee2010multichannel}.

In this paper, we propose to apply it for signature detection in a
multispectral image. We build upon our previous
work~\cite{rousseau2013compressive} that dealt with compressive
template detection and propose to extend it to perform compressive
pattern detection.

The outline of the paper is as follows. In
section~\ref{sec:template-matching}, we introduce the template
matching minimization problem of Guo and Osher and we bring elements
of a response that show why that minimization problem is succeeding.
We then briefly present how we solve this minimization problem and its
variants. In section~\ref{sec:compr-templ-match}, we explain how this
minimization problem can be extended to work with compressive data and
we give a few numerical experiments demonstrating the performance of
the compressive template matching minimization problem. In
section~\ref{sec:compr-patt-detect}, we study how we can extend the
compressive template matching minimization problem to perform
compressive pattern matching and provide numerical experiments. We
give conclusions and perspectives in section~\ref{sec:conclusion}.

\section{Template matching}
\label{sec:template-matching}

\subsection{$L_1$-based template matching}
\label{sec:l_1-based-template}

Template detection in a multispectral image is one of the first
application when dealing with multispectral data. It consists of
locating a template within a multispectral image. With the constant
growing of the numbers of channels, this problem becomes
computationnaly challenging. Several algorithms have been proposed so
far to tackle down this problem, see~\cite{guo2011template} and
references therein. On a recent paper in~\cite{guo2011template}, the
following minimization problem is suggested:
\begin{equation}\label{eq:1}
  \argmin_{u\geq 0}\|u\|_1 \quad\textrm{s.t.}\quad \|\bX^Tu-s\|_2<\sigma.
\end{equation}
In this minimization problem, $\bX$ is a matrix that stores the data
collected by multispectral sensors. Each column corresponds to a
channel and each row is the spectrum of a pixel. The vector $s$ is the
template we want to detect. We give here intuitive arguments
explaining why this minimization problem is working. First, the term
$\bX^Tu$ can be interpreted as the linear combination of the rows of
$\bX$ weighted by the vector $u$. So $\bX^Tu$ is in fact a linear
combination of the pixels of $\bX$ weighted by $u$. Moreover, the
$L_1$ norm promotes sparse solutions. As a result, the minimization
problem is looking for a reduced set of pixels whose linear
combination with coefficients in $u$ yields $s$. There are three types
of solutions that satisfy that constraint:
\begin{itemize}
  \item The solutions where $u$ has a non-zero entry at every pixel of
  spectral signature $s$. That way, we would have a linear combination
  of spectral signatures approaching $s$ which would give $s$.
  \item The solutions where $u$ has a non-zero entry at some pixels of
  spectral signature $s$ but not all of them.
  \item The solutions that combine random pixels of $\bX$ and
  nonetheless yields $s$.
\end{itemize}

\begin{figure}[ht]
  \centering
  \begin{subfigure}[b]{.4\linewidth}
    \input{tikz/NormeL1_CS}
    \caption{Classic compressive sensing case}
    \label{fig:L1ball:CS}
  \end{subfigure}
  \quad
  \begin{subfigure}[b]{.4\linewidth}
    \input{tikz/NormeL1_Template}
    \caption{Solutions to $\bX^Tu=s$ minimizing the $L_1$ norm}
    \label{fig:L1ball:template}
  \end{subfigure}
  \caption{Comparison with the classic compressive sensing case}
  \label{fig:L1ball}
\end{figure}
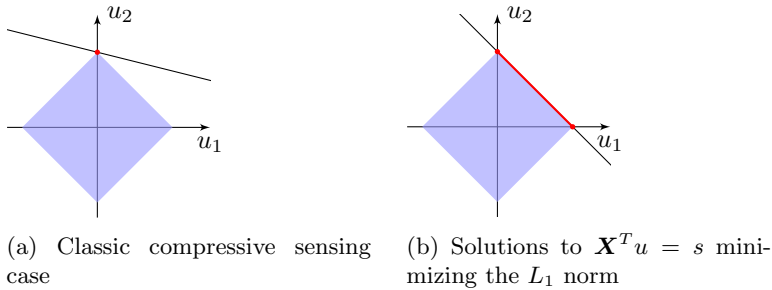

The latter type of solutions is ruled out because the $L_1$ norm of
the corresponding $u$ is likely to be greater than 1. Solutions of the
first two types are then preferred. Figure~\ref{fig:L1ball} helps us
to understand why solutions of the first type are preferred. In
Fig.~\ref{fig:L1ball:CS}, we display the classic compressive sensing
case where the $L_1$ minimization problem do find a sparse solution.
Figure~\ref{fig:L1ball:template} illustrates the case we are
interested in where there are infinitely many solutions that minimize
the $L_1$ norm and verify the constraint which are displayed in red.
Some of the solutions are sparse (the two red points on the axis in
our case) and represent solutions of the second type. The other
solutions are less sparse but have the same $L_1$ norm and we observe
that algorithms tends to select those solutions that have the largest
support among those that minimize the $L_1$ norm. That is why
solutions of the first type are preferred.

\subsection{Numerical solution of the minimization problem}
\label{sec:implementation}

In the following sections, we will have to solve minimization problems
of the form~\eqref{eq:1} and to improve the results, we will add a
regularization term based on total variation. We then have the
following minimization problem
\begin{equation}
  \label{eq:30}
  \argmin_{u\geq 0}\|u\|_1+\norm{u}_{\text{TV}}
  \quad\text{s.t.}\quad
  \|\bA u-f\|_2<err,
\end{equation}
Those minimizations are special cases of the more general minimization
problem
\begin{equation}
  \label{eq:23}
  \argmin_{u\geq 0}\|\bphi(u)\|_1\quad\text{s.t.}\quad \|\bA u-f\|_2<err,
\end{equation}
where $\bphi$ is a linear operator that yields model~\eqref{eq:1} when
equal to the identity or model~\eqref{eq:30} when equal to
\begin{equation}
  \label{eq:29}
  \begin{pmatrix}
    \Id\\
    \bD_x\\
    \bD_y
  \end{pmatrix},
\end{equation}
where $\Id$ is the identity matrix and $\bD_x$ and $\bD_y$ are
respectively the linear operators giving the gradient along $x$ and
$y$.

Due to their simplicity and flexibility, we use split Bregman
algorithms~\cite{goldstein2009split,cai2009split} to solve these
minimizations. We first apply the Bregman iteration to~\eqref{eq:23}
which gives
\begin{subnumcases}{\label{eq:24}}
  \label{eq:25}
  u^{k+1}=\argmin_{u\geq0}\norm{\bphi(u)}_1+\frac{\beta_1} 2\|\bA u-f^k\|^2_2\\
  f^{k+1}=f^k+f-\bA u^{k+1}
\end{subnumcases}
with $f_0=f$. Then, we use a splitting technique
(see~\cite{wang2007fast,goldstein2009split}) that introduces a new
unknown to solve each subproblem~\eqref{eq:25}.
\begin{equation}
  \argmin_{u\geq0, d}\|d\|_1+\frac{\beta_1} 2\|\bA u-f^k\|^2_2
  \quad\textrm{s.t.}\quad
  d=\bphi(u).
\end{equation}
We then apply the Bregman iteration once more because we now have a
constrained problem. We solve the minimization problem over $u$ and
$d$ with an alternating minimization by first minimizing with respect
to $u$ and then with respect to $d$.
\begin{subnumcases}{\label{equ:BSB:SB}}
  \begin{multlined}[c]
    u^{l+1}=\argmin_{u\geq0}\frac{\beta_1}2\|\bA u-f^k\|^2_2\\
    +\frac{\beta_2} 2\|d^l-\bphi(u)-b^l\|^2_2
  \end{multlined}
  \\
  d^{l+1}=\argmin_d \|d\|_1+\frac{\beta_2} 2\|d-\bphi(u^{l+1})-b^l\|^2_2\\
  b^{l+1}=b^l+\bphi(u^{l+1})-d^{l+1}
\end{subnumcases}
\begin{algorithm}[h]
  \SetKwInOut{KwSolve}{Solve}
  \KwSolve{$\displaystyle\argmin_{u\geq
      0}\|\bphi(u)\|_1\quad\text{s.t.}\quad \|\bA u-f\|_2<err$}
  \KwData{$\bphi, \bA, f, \beta_1, \beta_2, err$}
  \KwResult{}
  $f^0:=f$\;
  $b^0:=0$\;
  $d^0:=0$\;
  $k:=0$\;
  $\bD_{\mathrm{inv}}:=(\beta_1 \bA^T\bA+\beta_2\bphi^T\bphi)^{-1}$\;
  \Repeat{$\|\bA u^{k+1}-f\|_2<err$}{
    $u^{k+1}:=\bD_{\mathrm{inv}}(\beta_1\bA^T(f-f^k)+\beta_2\bphi^T(d^k-b^k))$\;
    $u^{k+1}:=\max(u^{k+1}, 0)$\;
    $d^{k+1}:=\shrink(\bphi(u^{k+1})+b^k, \beta_2)$\;
    $b^{k+1}:=b^k+\bphi(u^{k+1})-d^{k+1}$\;
    $f^{k+1}:=f^k+f-\bA u^{k+1}$\;
    $k:=k+1$\;
  }
  \caption{Constrained Split Bregman}
  \label{alg:splitbregman}
\end{algorithm}
If we limit the number of iterations to solve $u^{k+1}$ to only one in
equations~\eqref{equ:BSB:SB} and plug it back in
equations~\eqref{eq:24}, we finally have
\begin{subnumcases}{}
  \label{eq:26}
  \begin{multlined}[c]
    u^{k+1}=\argmin_{u\geq0}\frac{\beta_1}2\|\bA u-f^k\|^2_2\\
    +\frac{\beta_2} 2\|d^k-\bphi(u)-b^k\|^2_2
  \end{multlined}\\
  \label{eq:27}
  d^{k+1}=\argmin_d \|d\|_1+\frac{\beta_2} 2\|d-\bphi(u^{k+1})-b^k\|^2_2\\
  b^{k+1}=b^k+\bphi(u^{k+1})-d^{k+1}\\
  f^{k+1}=f^k+f-\bA u^{k+1}
\end{subnumcases}
The minimization problem~\eqref{eq:26} is solved by first dropping the
constraint $u\geq0$, solving the resulting least square problem and
then forcing the solution to be non-negative. The minimization
problem~\eqref{eq:27} admits a closed-form solution using the shrinkage
operator $\shrink$ defined as follows
\begin{equation}
  \label{eq:28}
  \shrink(x, \lambda)=\sgn(x)\bullet\max(0, \abs{x}-\frac{1}{\lambda}),
\end{equation}
where the operator $\bullet$ is the entry-wise product.

The detailed algorithm is shown in Algorithm~\ref{alg:splitbregman}.
In our experiments, we take $\bA=\bX^T$, $f=s$, $err=10^{-2}$,
$\beta_1=1$ and $\beta_2=1000$.

\section{Compressive template matching}
\label{sec:compr-templ-match}

The previous template matching minimization problem needs the whole
multispectral data cube to work. In this section, we propose a new
minimization problem working in a compressive sensing context where we
only have access to a small number of linear measurements on the
multispectral image. The idea of processing signals without
reconstructing them first dates back to the work of Davenport
\textit{et al}~\cite{davenport2006detection}. That bypassing idea has
then been applied to multispectral image processing problems such as
unmixing. Here, we wish to reconstruct the vector $u$ solution
of~\eqref{eq:1} without reconstructing the image first. This section
and the following ones extends the work done in a previous article
~\cite{rousseau2013compressive}.

\subsection{Measurement model and problem formulation}
\label{sec:meas-model-probl}

We recall that the multispectral image is stored in a matrix $\bX$
where each column corresponds to a channel that is vectorized and each
row is the spectrum of a pixel. We will assume that the multispectral
image has $n_P$ pixels and $n_B$ bands. As a result, the matrix $X$
has $n_P$ rows and $n_B$ columns. The acquisition model is described
as
\begin{equation}
  \bM=\bF\bX,
\end{equation}
where $\bF$ is a sensing matrix. This amounts to perform independently
the same measurements on each band of the image and store them is the
corresponding column of $\bM$. The sensing matrix $F$ has to verify of
few properties for the recovering minimization problem to work. A
popular one is the RIP condition that is known to be verified by
independent and identically distributed (iid) Gaussian sensing matrix.
However, these matrices are physically unrealistic and one considers
simpler matrices such as Bernouilli distributed ones or Hadamard
matrix. In this paper, we keep using Gaussian distributed sensing
matrices as a reference but also use Gaussian distributed circulant
matrices that are much more realistic from a physical point of view
and that still are good sensing matrix~\cite{romberg2009compressive}.

We then define the measurement rate $p$ where $0<p<1$ (also expressed
as a percentage), and what is its influence on the size of $\bF$. The
measurement rate $p$ is the fraction of the overall data that we want
to acquire. It means that the number of elements of $\bM$ is the number
of elements of $\bX$ multiplied by $p$. If the matrix $\bF$ is of size
$m\times n_P$ then the sensing matrix $\bM$ is of size $m\times n_B$
and we have the relation
\begin{equation}
  m\cdot n_B=p\cdot n_B\cdot n_P,
\end{equation}
which gives
\begin{equation}\label{eq:17}
  m = p\cdot n_P.
\end{equation}
The number of rows of $\bF$ is an integer so we will take $m=\lfloor
p\cdot n_P\rfloor$ where $\lfloor\cdot\rfloor$ is the operator mapping
a number to its largest previous integer.

The problem is then to solve the following minimization
\begin{equation}\label{eq:3}
  \argmin_{u\geq 0}\|u\|_1 \quad\textrm{s.t.}\quad
  \begin{cases}
    \|\bX^Tu-s\|_2<\sigma\\
    \bM=\bF\bX
  \end{cases},
\end{equation}
where we add the constraint coming from the measurements. The problem
is to eliminate $\bX$ from those two constraints since we no longer
have access to the multispectral data.

\subsection{Compressive template minimization}
\label{sec:compr-templ-minim}

One way to eliminate $\bX$ from the two constraints $\bX^Tu=s$ and
$\bM=\bF\bX$ in~\eqref{eq:3} is to introduce a matrix between $\bX^T$
and $u$ of the form $\bF^T\bA$ so we could replace $\bX^T\bF^T$ by
$\bM^T$ and eliminate $X$. This matrix should theoretically be equal
to the identity. However that is impossible because $\bF^T\bA$ is not
invertible. Given a matrix $\bF$, we have to find a matrix $\bA$ such
that $\bF^T\bA\approx \bI_{n_P}$. In the following, we will consider
two candidates for $\bA$. The first candidate comes from the
observation that if $\bF$ is a Gaussian distributed matrix, we have
$\frac 1 m \bF^T\bF\approx \bI_{n_P}$ as showed
in~\cite{rudelson1999random}. We can then take $\bA=\frac 1 m \bF$ and
we will refer to this type of matrix as type 1 (T1).

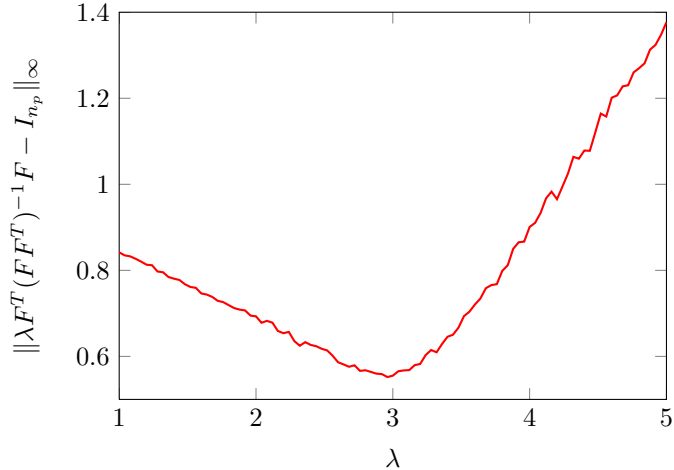
\begin{figure}
  \centering
  \input{gen/genlambda}
  \caption{$\lambda$ minimizing maximum error}
  \label{fig:lambda}
\end{figure}

One other candidate for $\bA$ is obtained by solving the following
minimization
\begin{equation}
  \argmin_{\bA}\norm{\bF^T\bA-\bI_{n_P}}_F,
\end{equation}
where $\norm{\cdot}_F$ is the Frobenius norm which is basically the
Euclidean norm of the vectorized matrix. This is a well know problem
involving the pseudo-inverse of $\bA$. One can show that the solution
writes $\bA=(\bF^T)^+$ where the $+$ operator is the pseudo-inverse.
Given that $\bF$ is a sensing matrix, we will always suppose that it
is of full rank. In that case, $\bA$ has an explicit formulation,
$\bA=(\bF\bF^T)^{-1}\bF$. However, that minimization does not help us
determining $\bF$. In fact, we can show that if $\bF$ is of full rank,
the norm $\norm{\bF^T(\bF\bF^T)^{-1}\bF-\bI_{n_p}}_F$ is constant and
is equal to $\sqrt{n_P-m}$. Among all matrices of full rank $\bF$, some
are obviously better than others for a sensing matrix. For example,
the matrix
\begin{equation}
  \bF=
  \begin{pmatrix}
    \bI_m&0
  \end{pmatrix},
\end{equation}
is a very bad candidate because we have
\begin{equation}
  \bF^T(\bF\bF^T)^{-1}\bF=
  \begin{pmatrix}
    \bI_m&0\\
    0&0
  \end{pmatrix},
\end{equation}
We note that the distance to $\bI_{n_p}$ is concentrated in a few
entries which is why this is a terrible choice for a sensing matrix.
Rather, we would like the error to be equally shared between all the
entries of $\bF$. We then choose the max norm instead of the $L_2$
norm. Keeping $(\bF\bF^T)^{-1}\bF$ as a possible candidate, we are now
interested in the minimization
\begin{equation}
  \argmin_\lambda\|\lambda \bF^T(\bF\bF^T)^{-1}\bF-\bI_{n_P}\|_\infty,
\end{equation}
where $\norm\cdot_\infty$ denotes the maximum norm. The solution is
the right scaling of the candidate $(\bF\bF^T)^{-1}\bF$ so as to minimize
the maximum error. If $\bF$ is iid Gaussian, we have already seen that
$\bF\bF^T\approx n_P\bI_m$. We have also $\bF^T\bF\approx m\bI_{n_P}$. This
suggests that $\lambda=\frac{n_P}{m}$. This is indeed what we find in
Fig.~\ref{fig:lambda} where $m=30$ and $n_P=100$, the minimum error is
at $\lambda\approx 3\approx\frac{n_P}m$.

\begin{figure}[ht]
  \centering
  \input{gen/generrmax}
  \caption{Maximum error}
  \label{fig:norm}
\end{figure}
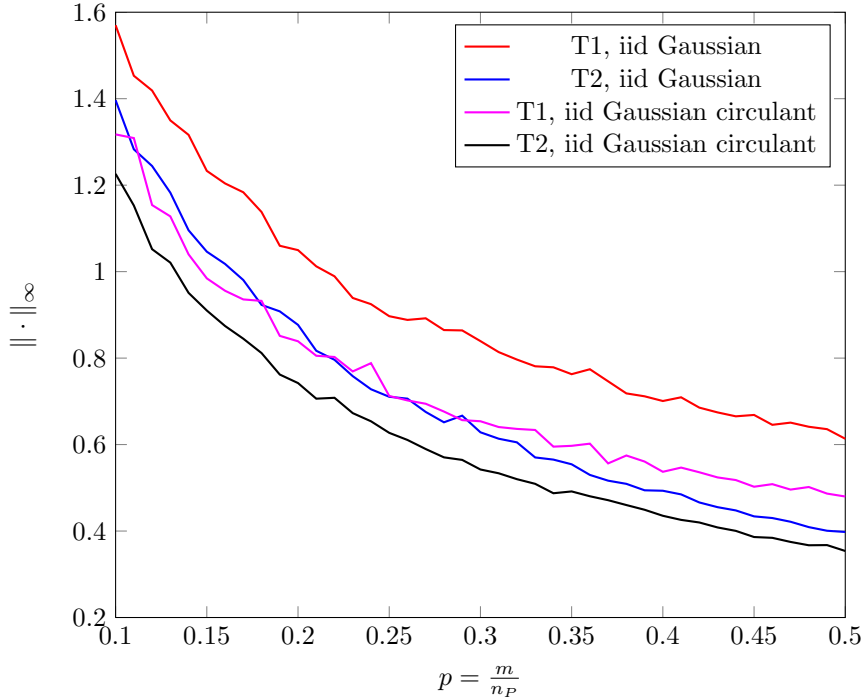

We have two candidates. Type 1 (T1) is $\frac 1 m\bF$ and type 2 (T2)
is $\frac{n_P}{m}(\bF\bF^T)^{-1}\bF$. Figure~\ref{fig:norm} depicts the
maximum error for the two types of matrix $\bA$. We also try two types
of sensing matrix $\bF$: iid Gaussian and iid Gaussian circulant.
Circulant matrices are used as sensing matrices because it has been shown
to be almost as effective as the Gaussian random matrix for CS
encoding/decoding~\cite{yin2010practical,romberg2009compressive}. Even
if candidate of type 1 come from a minimization of the Frobenius norm,
it is actually performing better than type 2. Quite surprisingly, the
smallest error is obtained when $\bF$ is the first $m$ rows of a
circulant matrix generated from a iid Gaussian vector. An intuitive
explanation of this is that choosing a reduced set of Gaussian
coefficients for a circulant matrix (the first line only) rather than
a whole matrix reduces the chance of hitting a large number in
absolute value that would give a large inner product (an entry in
$\bF\bF^T$ or $\bF^T(\bF\bF^T)^{-1}\bF$). This large entry is then the
final error since we are calculating the max norm.

From now on, we will consider matrix $\bA$ of type 2. We can now
eliminate $\bX$ from $\bX^Tu=s$ and $\bM=\bF\bX$. We have
\begin{equation}
  \bX^T\left(\frac{n_P}{m}\bF^T(\bF\bF^T)^{-1}\right)\bF u=s,
\end{equation}
and then, using $\bM=\bF\bX$, we have
\begin{equation}
  \bM^T(\bF\bF^T)^{-1}\bF u=\frac m {n_P} s.
\end{equation}
The minimization problem then becomes
\begin{equation}\label{eq:19}
  \argmin_{u\geq 0}\|u\|_1
  \quad\textrm{s.t.} \quad
  \|\bM^T(\bF\bF^T)^{-1}\bF u-\frac m {n_P}s\|_2<\sigma',
\end{equation}
with $\sigma'=\frac{n_P}{m}\sigma$.

Finally, this minimization problem is a modified version
of~\eqref{eq:1} but instead of using $\bX^T$ we use the matrix
$\bM^T(\bF\bF^T)^{-1}\bF$ that is an approximation of $\bX^T$.

\subsection{Numerical experiments}
\label{sec:numer-exper-1}

In this section, we illustrate both the regular template matching
minimization problem and its compressive counterpart onto two images.
We test both minimization problems with a measurement rate varying
from 1 to 40 percent. In both cases, once the algorithm finished, we
apply the Lloyd-Max clustering algorithm~\cite{max1960quantizing} on
the recovered $u$ to decide whether the detection is positive or not
for each pixel. We then count the number of errors by comparing the
resulting mask with the desired mask result. This reference mask is
calculated manually by choosing all the pixels that have a signature
approaching the one we want to detect. To improve the readability of
the results, the images are inverted before display. The algorithm
runs in less than a minute on a classic computer.

\begin{figure}[ht]
  \centering
  \def\scale{1.5}
  \begin{subfigure}[b]{.4\linewidth}
    \centering
    \includegraphics[scale=\scale]{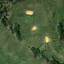}
    \caption{Original Giza color image}
    \label{Giza.Bunker.Template.orig.png}
  \end{subfigure}
  \quad
  \begin{subfigure}[b]{.4\linewidth}
    \centering
    {%
      \setlength{\fboxsep}{0pt}%
      \fbox{\includegraphics[scale=\scale]{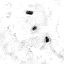}}
    }
    \caption{Template matching on Fig.~\subref{Giza.Bunker.Template.orig.png}}
    \label{Giza.Bunker.Template.X_L1.png}
  \end{subfigure}\\
  \begin{subfigure}[b]{.4\linewidth}
    \centering
    {%
      \setlength{\fboxsep}{0pt}%
      \fbox{\includegraphics[scale=\scale]{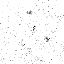}}
    }
    \caption{Compressive template matching, 30\% of data, L1
      regularizer}
    \label{Giza.Bunker.Template.L1_circ.p30.png}
  \end{subfigure}
  \quad
  \begin{subfigure}[b]{.4\linewidth}
    \centering
    {%
      \setlength{\fboxsep}{0pt}%
      \fbox{\includegraphics[scale=\scale]{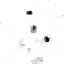}}
    }
    \caption{Compressive template matching, 30\% of data, TV/L1
      regularizer}
    \label{Giza.Bunker.Template.TVL1_circ.p30.png}
  \end{subfigure}
  \caption{Signature detection on the Giza image}
  \label{fig:giza.template}
\end{figure}

The first image is a $64\times 64$ color image of Giza,
Egypt\footnote{Available at
  \url{http://opticks.org/confluence/display/opticks/Sample+Data#SampleData}}
displayed in Fig.~\ref{Giza.Bunker.Template.orig.png}. The spectral
signature $s$ we want to detect is extracted from sandy areas. The
result of the template matching minimization~\eqref{eq:1} is shown in
Fig.~\ref{Giza.Bunker.Template.X_L1.png}. The shape of all three sandy
areas is well recovered.
Figure~\ref{Giza.Bunker.Template.L1_circ.p30.png} shows the
compressive template matching~\eqref{eq:19} with a $L_1$ regularizer
for a measurement rate of 30\%. We see that the detected pixels are
scattered in the image. We improve the detection by mixing the $L_1$
regularizer with a geometrical one as we can see in
Fig.~\ref{Giza.Bunker.Template.TVL1_circ.p30.png} where a $TV/L_1$
regularizer is used.

\begin{figure}
  \centering
  \input{gen/template_giza/template_giza_plot.tex}
  \caption{Percentage of wrong detection for different measurement
    rate, sensing matrix and regularizer on the Giza image}
  \label{fig:plot_error}
\end{figure}
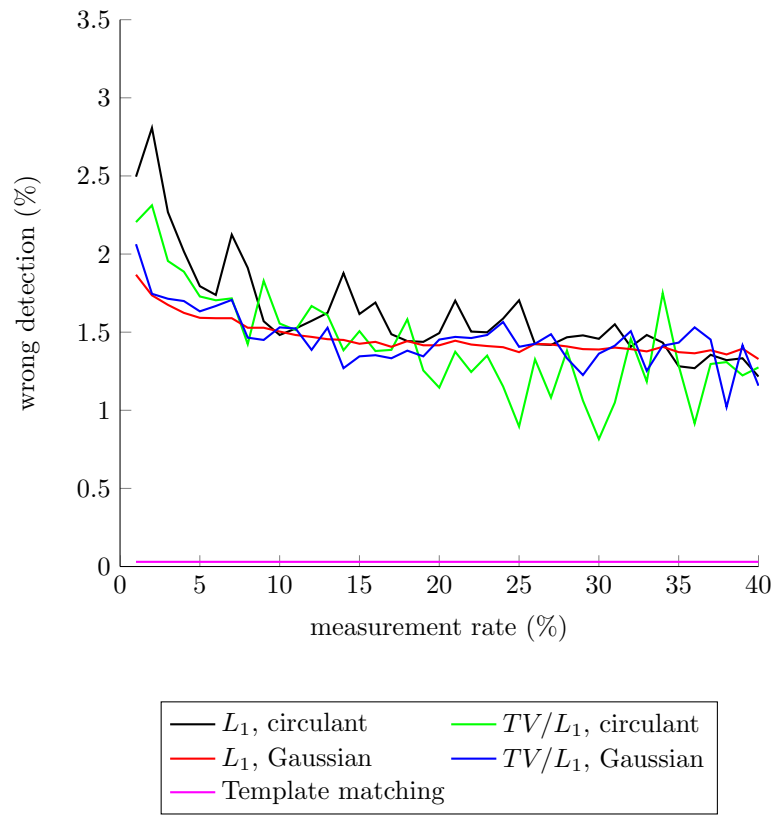

Figure~\ref{fig:plot_error} shows the results when the measurement
rate is varying. As a reference, the percentage of wrong detection of
the template matching minimization~\eqref{eq:1} is 0.03 \%. The
minimization performs best when the sensing matrix is Gaussian and the
regularizer is $TV/L_1$. Increasing the measurement rate after 10\%
does not improve much the detection. The more realistic case where the
sensing matrix is circulant is doing quite well when the measurement
rate is under 10\%. Again, increasing the measurement rate after 10\%
does more harm than good. On the contrary, the $L_1$ regularizer shows
a constantly decreasing error rate as the measurement rate increases.
Another indicator that shows that Gaussian circulant sensing matrices
are good sensing matrices is that pure Gaussian sensing matrices perform
only slightly better than Gaussian circulant sensing matrices.

\begin{figure}[ht]
  \centering
  \input{gen/moffett_signature}
  \caption{Spectral signature of buildings}
  \label{Moffett.Signature}
\end{figure}
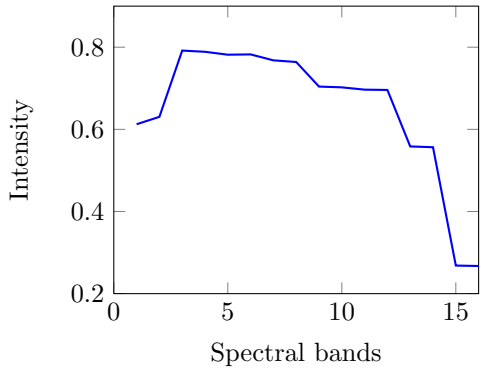

\begin{figure}[ht]
  \centering
  \def\scale{1.5}
  \begin{subfigure}[b]{.4\linewidth}
    \centering
    \includegraphics[scale=\scale]{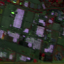}
    \caption{Moffett field image in false color}
    \label{Mof.Fac.Template.orig}
  \end{subfigure}
  \quad
  \begin{subfigure}[b]{.4\linewidth}
    \centering
    {%
      \setlength{\fboxsep}{0pt}%
      \fbox{\includegraphics[scale=\scale]{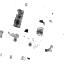}}
    }
    \caption{Template matching on
      Fig.~\subref{Mof.Fac.Template.orig}}
    \label{Mof.Fac.Template.X_L1}
  \end{subfigure}\\
  \begin{subfigure}[b]{.4\linewidth}
    \centering
    {%
      \setlength{\fboxsep}{0pt}%
      \fbox{\includegraphics[scale=\scale]{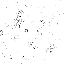}}
    }
    \caption{Compressive template matching, 30\% of data, $L_1$
      regularizer}
    \label{Mof.Fac.Template.L1.p30}
  \end{subfigure}
  \quad
  \begin{subfigure}[b]{.4\linewidth}
    \centering
    {%
      \setlength{\fboxsep}{0pt}%
      \fbox{\includegraphics[scale=\scale]{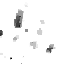}}
    }
    \caption{Compressive template matching, 30\% of data, $TV/L_1$
      regularizer}
    \label{Mof.Fac.Template.TVL1.p30}
  \end{subfigure}
  \caption{Signature detection on Moffett field image}
  \label{fig:moffet.template}
\end{figure}

The second multispectral image is extracted from the Moffett Field
AVIRIS multispectral image\footnote{Available at
  \url{http://aviris.jpl.nasa.gov/html/aviris.freedata.html}}. We
selected 16 bands out of the 224 available and extracted a $64\times
64$ image of interest shown in Fig.~\ref{Mof.Fac.Template.orig}. We
would like to detect the spectral signature of buildings shown in
Fig.~\ref{Moffett.Signature}. Figure~\ref{fig:moffet.template} shows
some of the results. As a reference, the percentage of wrong detection
of the template matching minimization~\eqref{eq:1} is 0.3 \%.

\begin{figure}
  \centering
  \input{gen/signature_moffet/signature_moffett_plot}
  \caption{Percentage of wrong detection for different measurement
    rate, sensing matrix and regularizer on the Moffett image}
  \label{fig:plot_error_moffett}
\end{figure}
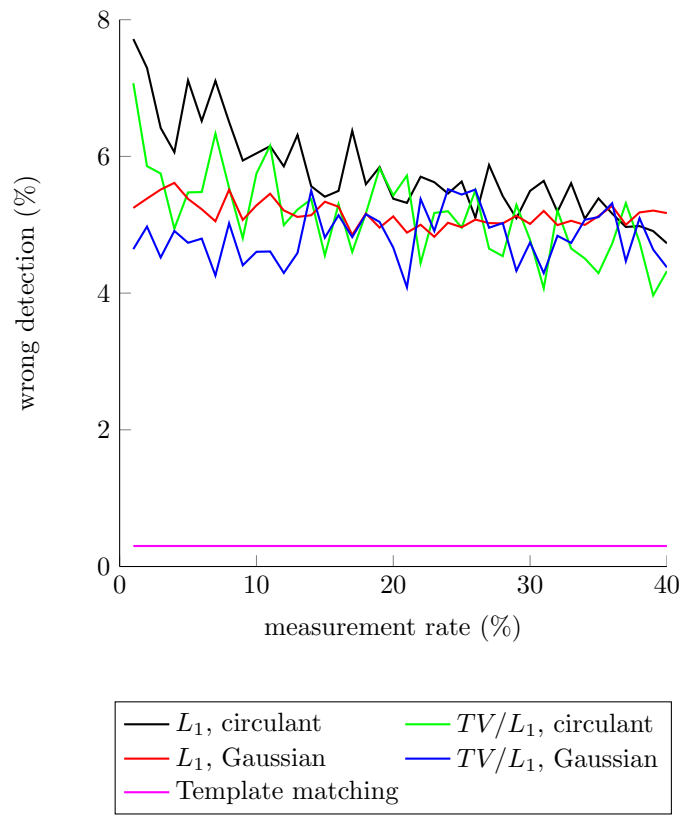

Again, according to Fig.~\ref{fig:plot_error_moffett}, the compressive
template minimization performs best when the sensing matrix is
Gaussian and the regularizer is $TV/L_1$. However, the difference is
less obvious than in the previous experiment but we can still see that
$TV/L_1$-based results show a better detection of connected objects.




\section{Compressive pattern detection}
\label{sec:compr-patt-detect}

In the previous section, we developed a minimization problem to detect
the location of a known spectral signature from a limited number of
measurements without reconstructing the image. The purpose of this
section is to generalize that minimization problem to deal with
pattern detection. We first introduce a formal transform on a
multispectral image called \textit{spectralization} that depends on
the pattern $P$ we wish to find. That formal transform essentially
adds the pattern information as spectral information in the new image.
Then, we use the previous minimization problem to detect signatures in
this new image which are in fact patterns in the original image. As we
will see, we need to reconstruct measurements because in fact we only
have measurements on the original image and not on the
\textit{spectralized} one.

\subsection{Image \textit{spectralization} along a pattern}

We first have to define the structure of the pattern we wish to detect
before proceeding. It can be modelled as a finite subset of $\mathbb
Z^2$. For example, if the structure we are interested in is a $2\times
2$ hook, the corresponding pattern as a subset of $\mathbb Z^2$ will
be represented as
\begin{equation}\label{eq:31}
  \pattern{(0, 0), (1, 0), (1, 1)}
\end{equation}
We then need to fix an order on that subset. The first point which
will serve as a reference point is always the point $(0, 0)$. The
pattern's structure is now a $n$-tuple of couples. In the previous
example, we choose
\begin{equation}
  P=((0, 0), (1, 0), (1, 1)).
\end{equation}
Now that we have an ordered structure of the pattern $P$, we can
define the \textit{spectralization} of an image $\bI$ with respect to
$P$. Roughly speaking, the \textit{spectralization} of $\bI$ is just
stacked copies of $\bI$ that are shifted according to the ordered
pattern we choose. This can be generalized to multispectral images.
Let us first introduce some definitions and convenient notations
related to image shifting and matrix stacking. We will then define the
\textit{spectralized} image of a grayscale image and extend it to
multispectral images.

Given an element $p=(i, j)$ of $\mathbb Z^2$, we define the operator
$S_{p}$ on a matrix which shifts all its entries by $p$. For example,
the entry at $(1, 1)$ is shifted to the place $(i+1, j+1)$. More
formally, if $\bI$ is of size $N\times P$ we have
\begin{equation}
  [S_{p}(\bI)]_{ab}= \bI_{(a+i-1 \mod N) + 1, (b+j-1 \mod P) + 1}.
\end{equation}
We have the obvious properties that $S_{(0, 0)}$ is the identity,
$S_{(0, 0)}=\id$ and that composing $S_{p_1}$ and $S_{p_2}$ yields
$S_{p_1+p_2}$, $S_{p_1+p_2}=S_{p_1}\circ S_{p_2}$. If the pattern's
structure is
\begin{equation}
  P=(p_1,\dots,p_{|P|}),
\end{equation}
with the convention that $p_1=(0, 0)$, we will consider the $|P|$
corresponding shifting functions $S_{p_1}=\id, S_{p_2},\dots,
S_{p_{|P|}}$ where $\abs P$ denotes the number of elements of $P$.

The operator $S_{p}$ is acting on $\bI$; the corresponding matrix
acting on $\vecc(\bI)$, the vectorized version of the matrix $I$, is
denoted $\bQ_p$ and we have
\begin{equation}\label{eq:2}
  \vecc(S_p(\bI))=\bQ_p(\vecc(\bI)).
\end{equation}
Since the matrix $\bQ_p$ is a permutation matrix, it is orthogonal and
we have
\begin{equation}
  \label{eq:15}
  \bQ_{-p}=\bQ_p^{-1}=\bQ_p^T.
\end{equation}

We will also need to stack and merge matrices. If $\bA$ and $\bB$ are
matrices with the same number of rows, $[\bA, \bB]$ will denote the
concatenation of $\bA$ and $\bB$ along their columns. If $\bA$ and
$\bB$ are two matrices of the same size, $\bA\wedge \bB$ will denote
the 3-dimensional matrix obtained by stacking them along a third
dimension.

We can now give a precise definition of the \textit{spectralization}
of the matrix $\bI$ with respect to the pattern
$P=(p_1,\dots,p_{|P|})$. It is a 3-dimensional matrix obtained by
stacking the matrices $S_{-p_1}(\bI),\dots,S_{-p_{|P|}}(\bI)$. For
example, if the image $\bI$ and the pattern $P$ are defined as follows
\def\onelabel{\label{eq:32}}
\ifCLASSOPTIONonecolumn
\begin{equation}\onelabel
  \bI=
  \begin{pmatrix}
    1&2&3\\
    4&5&6\\
    7&8&9
  \end{pmatrix}
  \quad\text{and}\quad
  P=((0, 0), (1, 0), (1, 1))\quad\text{or}\quad\pattern{(0, 0), (1, 0), (1, 1)},
\end{equation}
we have
\begin{equation}
  S_{(0, 0)}(\bI)=\bI,\quad S_{-(1, 0)}(\bI)=
  \begin{pmatrix}
    4&5&6\\
    7&8&9\\
    1&2&3
  \end{pmatrix},\quad S_{-(1, 1)}(\bI)=
  \begin{pmatrix}
    5&6&4\\
    8&9&7\\
    2&3&1
  \end{pmatrix}.
\end{equation}
\else
\begin{gather}
  \bI=
  \begin{pmatrix}
    1&2&3\\
    4&5&6\\
    7&8&9
  \end{pmatrix},\\
  \onelabel
  P=((0, 0), (1, 0), (1, 1))\quad\text{or}\quad\pattern{(0, 0), (1, 0), (1, 1)},
\end{gather}
we have
\begin{align}
  S_{(0, 0)}(\bI)&=
  \begin{pmatrix}
    1&2&3\\
    4&5&6\\
    7&8&9
  \end{pmatrix}=\bI,\\
  S_{-(1, 0)}(\bI)&=
  \begin{pmatrix}
    4&5&6\\
    7&8&9\\
    1&2&3
  \end{pmatrix},\\
  S_{-(1, 1)}(\bI)&=
  \begin{pmatrix}
    5&6&4\\
    8&9&7\\
    2&3&1
  \end{pmatrix}.
\end{align}\fi
The \textit{spectralized} image $\spec_P(\bI)$ of $\bI$ with respect
to the pattern's structure $P$ is the 3-dimensional matrix $[\bI
\wedge S_{-(1, 0)}(\bI) \wedge S_{-(1, 1)}(\bI)]$. In its linearized
form, we have $\spec_P(\bI)=[\vecc(\bI), \vecc(S_{-(1, 0)}(\bI)),
\vecc(S_{-(1, 1)}(\bI))]$ which can also be written $[\vecc(\bI),
\bQ_{-(1, 0)}\vecc(\bI), \bQ_{-(1,1)}\vecc(\bI)]$ using~\eqref{eq:2}.
We then have the following definition.
\begin{defn}
  Given a gray-scale image $\bI$ and a pattern structure $P$, the \textit{spectralized}
  image $\spec_P(\bI)$ is the matrix
  \begin{equation}
    [\vecc(\bI), \bQ_{-p_2}\vecc(\bI),\dots,\bQ_{-p_{|P|}}\vecc(\bI)].
  \end{equation}
\end{defn}

Using the previous example, we have
\begin{equation}\label{eq:12}
  \spec_P(\bI)=
  \begin{pmatrix}
    1&4&5\\
    4&7&8\\
    7&1&2\\
    \mathbf{2}&\mathbf{5}&\mathbf{6}\\
    5&8&9\\
    8&2&3\\
    3&6&4\\
    6&9&7\\
    9&3&1
  \end{pmatrix}
\end{equation}
Thus, if we want to detect the pattern
\begin{equation}
  \begin{pmatrix}
    2&\star\\
    5&6
  \end{pmatrix},
\end{equation}
corresponding to the structure $P$ in the image $\bI$, it suffices to
look for the signature $(2, 5, 6)$ in the \textit{spectralized}
image~\eqref{eq:12}.

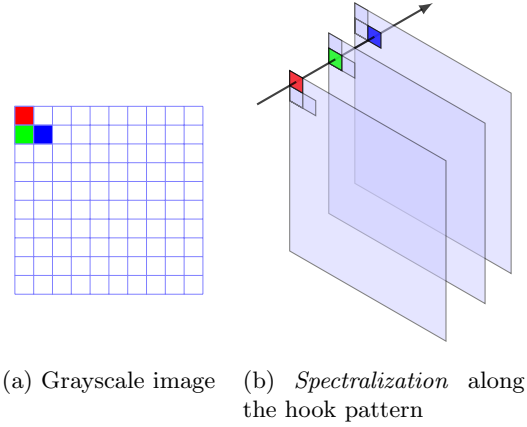
\begin{figure}[ht]
  \ffigbox
  {%
    \begin{subfloatrow}
      \ffigbox[\FBwidth][1.5\FBheight]
      {%
        \caption{Grayscale image}
        \label{tikz/GridInvader}
      }
      {%
        \tikzset{every picture/.style={scale=0.5}}%
        \begin{tikzpicture}[baseline=(current bounding box.center)]
          \begin{scope}[scale=.5, color=blue!50]
            \coordinate (O) at (0, 8);
            \draw[help lines, color=blue!50] (0, 0) grid (10, 10);
            \foreach \p/\color in {
              {(0, 1)}/red,
              {(0, 0)}/green,
              {(1, 0)}/blue} {
              \draw[fill=\color, shift=(O)] \p rectangle ++(1, 1);
            }
          \end{scope}
        \end{tikzpicture}
      }

      \ffigbox[\FBwidth]
      {%
        \caption{\textit{Spectralization} along the hook pattern}
        \label{tikz/CubeInvader}
      }
      {%
        \tikzset{every picture/.style={scale=0.5}}%
        \begin{tikzpicture}[baseline=(current bounding box.center)]
          \begin{scope}[scale=.4, z={ ($ 3*(0.866cm,0.5cm) $) }, x={(0.866cm,-0.5cm)},
            y={(0cm,-1cm)}]
            \coordinate (O) at (0, 0, 0);

            \draw[thick, -latex, opacity=.75] (.5, .5, -4) -- (.5, .5, .5);
            \foreach \refp/\color [count=\count] in {
              {(-1, -1)}/blue,
              {(-0, -1)}/green,
              {(-0, -0)}/red} {
              \begin{scope}[canvas is xy plane at z=-\count]
                \def\lenimage{12}
                \draw[fill=blue!20, opacity=.5] \refp rectangle +(\lenimage, \lenimage);

                \draw[fill=blue!20, opacity=.5] \refp
                rectangle ++(1, 2)
                rectangle ++(-1, -1)
                rectangle ++(2, 1);

                \draw[fill=\color, opacity=.75] (0, 0) rectangle (1, 1);

              \end{scope}
            }

            \draw[thick, opacity=.75] (0, 1, -3) -- (.5, .5, -3);
            \draw[thick, opacity=.75] (0, 1, -2) -- (.5, .5, -2);
            \draw[thick, opacity=.75] (-1, 2, -1) -- (.5, .5, -1);
          \end{scope}
        \end{tikzpicture}
      }

    \end{subfloatrow}}
  {%
    \caption{\textit{Spectralization} of a grayscale image along the
      hook pattern}
    \label{fig:spectralization.image}}
\end{figure}

Figure~\ref{fig:spectralization.image} shows the
\textit{spectralization} of a grayscale image. The pattern used is
shown in Fig.~\ref{tikz/GridInvader}. The \textit{spectralization} of
the image along this pattern is shown in Fig.~\ref{tikz/CubeInvader}.
As in the previous example, we see that the \textit{spectralized}
image is formed by copies of the original image put side by side. Each
of these copies is shifted.

Following the last writing of $\spec_P(\bI)$, we define the
\textit{spectralized} image of a multispectral image $\bX$ with
respect to $P$.
\begin{defn}
  Given a multispectral image $\bX$ and a pattern structure $P$, the
  \textit{spectralized} image $\spec_P(\bX)$ is the matrix,
  \begin{equation*}
    [\bX, \bQ_{-p_2}\bX,\dots, \bQ_{-p_{|P|}}\bX].
  \end{equation*}
\end{defn}

\subsection{Measurements reconstruction}

We have just established that a pattern detection on a multispectral
image $\bX$ is equivalent to a signature detection on the
\textit{spectralized} image of $\bX$ denoted by $\spec_P(\bX)$. The
problem is that we need measurements on $\spec_P(\bX)$ to solve the
minimization problem and the only measurements we have are on $\bX$.
We somehow have to reconstruct measurements on $\spec_P(\bX)$ based on
$\bX$'s ones.

To measure the efficiency of the reconstruction of measurements, we
introduce the following ratio depending on a given pattern $P$
\begin{equation}\label{eq:16}
  \alpha(P)=\frac{\text{effective measurements}}{\text{virtual
      measurements}}.
\end{equation}
The effective measurements are measurements performed on the real
image. These are measurements that we could use in our calculations.
On the contrary, virtual measurements are measurements on the fictive
image $\spec_P(\bX)$ that ought to be reconstructed from the effective
ones. The ratio $\alpha(P)$ measures how many more measurements we
need to make to fully reconstruct measurements on $\spec_P(\bX)$. The
ratio $\alpha(P)$ has simple bounds. Indeed, one virtual measurement
requires at least one effective measurement to be reconstructed. As a
result we have $\alpha(P)\geq 1$. On the other hand, one virtual
measurement requires at most $\abs{P}$ measurements on $\bX$ because
$\spec_P(\bX)$ has $\abs{P}$ times more bands than $\bX$. We then have
$\alpha(P)\leq\abs{P}$. In conclusion, we have
\begin{equation}
  1\leq\alpha(P)\leq|P|.
\end{equation}

For example, if we wish to detect the hook pattern consisting of 3
pixels, we have to make at most three times more measurements to
reconstruct all virtual measurements in order to apply the minimization.
If we want a measurement rate of 30\% of the virtual image we need at
most a 90\% measurement rate on the original image. Fortunately, this
upper bound can be lowered by properly choosing the measurements to
make.

\subsection{Reconstruction using shifted measurements}
\label{sec:reconstr-using-shift}

The first idea that comes to mind is to take measurements that are not
independent but shifted from one another. Indeed, a shifted
measurement on a shifted image could be the same as the original
measurement on the original image. One effective measurement could be
used to reconstruct more than one virtual measurement.

Let us first see how a measurement vector $f$ that is shifted by a
vector $e$ writes. Let $\bB$ be the measurement matrix that is the
2-dimensional version of the measurement vector $f$. We then have
\begin{equation*}
  \vecc(\bB)=f.
\end{equation*}
If the measurement matrix $\bB$ is shifted by a vector $e$, it becomes
$S_e(\bB)$. The vectorized form is $\vecc(S_e(\bB))$. According
to~\eqref{eq:2}, the shifted measurement vector $f_e$ is then
\begin{align}
  \nonumber f_e&=\vecc(S_e(\bB))\\
  \nonumber &=\bQ_e\vecc(\bB)\\
  &=\bQ_ef.
\end{align}

Suppose all our virtual measurements are a shifted version of one
measurement $f$. The shifting information is modelled as a subset $E$
of $\mathbb Z^2$ containing the point $(0, 0)$. The point $(0, 0)$
represents the original measurement. All the other points represent
shifted measurements of the original one. According to~\eqref{eq:15},
the measurements writes
\begin{align*}
  m_e&=(\bQ_ef)^T\spec_P(\bX)\\
  &=f^T\bQ_{-e}\spec_P(\bX),
\end{align*}
for all $e\in E$. This can be written $\bMvirt=\bFvirt\spec_P(\bX)$
where
\begin{equation*}
  E= \{e_1,\dots,e_{|E|}\},
\end{equation*}
and
\begin{equation}
  \bMvirt=
  \begin{pmatrix}
    m_{e_1}\\
    \vdots\\
    m_{e_{|E|}}
  \end{pmatrix}\quad\text{and}\quad
  \bFvirt=
  \begin{pmatrix}
    f^T\\
    f^T\bQ_{-e_2}\\
    \vdots\\
    f^T\bQ_{-e_{|E|}}
  \end{pmatrix}
\end{equation}
Further, we have
\begin{align}
  \nonumber\bMvirt&=
  \begin{pmatrix}
    f^T\\
    f^T\bQ_{-e_2}\\
    \vdots\\
    f^T\bQ_{-e_{|E|}}
  \end{pmatrix}
  \begin{bmatrix}
    \bX,& \bQ_{-p_2}\bX,&\dots,& \bQ_{-p_{|P|}}\bX
  \end{bmatrix}\\
  \nonumber&=\left(f^T\bQ_{-e_i}\bQ_{-p_j}\bX\right)_{\substack{1\leq i\leq |E|\\1\leq
      j\leq |P|}}\\
  \label{eq:9}
  &=\left((\bQ_{e_i+p_j}f)^T\bX\right)_{\substack{1\leq i\leq |E|\\1\leq
      j\leq |P|}}.
\end{align}
In other words, to reconstruct $\bMvirt$ that gathers the results of
measurements on the virtual image $\spec_P(\bX)$, we need to take the
shifted measurements represented by $E'=E+P$ where $E+P$ denotes the
set
\begin{equation}\label{eq:20}
\left\{e_i+p_j \mid e_i\in E, 1\leq i\leq |E|, p_j\in P, 1\leq j\leq
  |P|\right\}.
\end{equation}
Therefore, we choose
\begin{equation}\label{eq:21}
\bFeff=
\begin{pmatrix}
  f^T\\
  f^T\bQ_{-e'_2}\\
  \vdots\\
  f^T\bQ_{-e'_{|E'|}}
\end{pmatrix},
\end{equation}
as an effective sensing matrix, so we have
\begin{equation}\label{eq:22}
  \bMeff=\bFeff\bX=
  \begin{pmatrix}
    f^T\bX\\
    (\bQ_{e'_2}f)^T\bX\\
    \vdots\\
    (\bQ_{e'_{\abs{E'}}}f)^T\bX
  \end{pmatrix}.
\end{equation}
The matrix $\bMeff$ gathers the measurements taken on the real image
$\bX$ by the sensing matrix $\bFeff$ and it contains all the
information we need to reconstruct the matrix $\bMvirt$
in~\eqref{eq:9}.

We are now able to write the measurement reconstruction ratio
\begin{equation}\label{eq:14}
  \alpha(P)=\frac{|E+P|}{|E|},
\end{equation}
which reflects the fact that we have to take $\abs{E+P}$ effective
measurements to reconstruct the $\abs{E}$ virtual ones.

\begin{prop}\label{prop:x-2}
  For a fixed pattern $P$ we have
  \begin{equation*}
    \inf_{E}\frac{|E+P|}{|E|}=1.
  \end{equation*}
\end{prop}
\begin{proof}
  We already know that $|E+P|\geq|E|$ and hence $\frac{|E+P|}{|E|}\geq
  1$. Let $R$ be a $a$ rows and $b$ columns rectangle containing the
  pattern $P$. For $n\geq1$, let $E$ be a $na$ rows and $nb$ columns
  rectangle. It is easy to see that $E+R$ is a $(n+1)a-1$ rows and
  $(n+1)b-1$ columns rectangle. Thus we have
  \begin{equation*}
    \frac{|E+R|}{|E|}=\frac{\left((n+1)a-1\right)\left((n+1)b-1)\right)}{n^2ab}
    \tikz[baseline=-0.5ex]{\draw[->=latex', shorten >=3pt, shorten <=3pt] (0, 0) --
      node[below] (A) {\scriptsize $n\longrightarrow +\infty$} (2, 0);
    } 1.
  \end{equation*}
\end{proof}

This proposition suggests that we choose $E$ so that $|E|$ has the
highest possible value. Unfortunately, we have $|E|\leq|E+P|$ and
the number of effective measurements $|E+P|$ is limited by
the size of the image: we cannot take too many measurements.

The problem is then to find the structure of $E$ minimizing the ratio
with $|E|$ bounded. For a fixed pattern $P$, the problem is
\begin{equation}\label{eq:5}
  \argmin_{|E|= A}\frac{|E+P|}{|E|}.
\end{equation}

The problem~\eqref{eq:5} is a difficult one due to its combinatorial
nature. Still, we can solve it if the pattern $P$ has a simple shape.
Suppose that $P$ is a rectangle. To minimize $\alpha(P)$, the
measurement pattern $E$ should also be a rectangle that has the same
shape as the pattern $P$. This result is formalized in the following
proposition. The details of the proof are shown in appendix.
\begin{prop}\label{prop:1}
  Suppose that the pattern $P$ is a rectangle. Then, a measurement
  pattern $E$ such that $\abs E=A$ minimizing the ratio~\eqref{eq:14}
  is obtained when $E$ is rectangular-shaped and its height $h$
  minimizes the functional
  \begin{equation*}
    (a-1)\ceil{\frac A h}+(b-1)h,
  \end{equation*}
  where $a$ and $b$ are respectively the height and the width of $P$
  and $\ceil\cdot$ is the operator mapping a number to its smallest
  following integer.
\end{prop}

For example, suppose we wish to detect a rectangular pattern $P$ of
size $6\times 10$ in a $128\times 128$ color image. The
\textit{spectralized} image is then of size $128\times128$ and has
$6\times10\times3=180$ bands. Suppose we want a 25\% virtual
measurement rate, we then need $128\times128\times25/100=4096$
different measurement vectors according to~\eqref{eq:17}. Given that we
use shifted measurements, if $E$ denotes the shifting pattern of the
virtual measurements, we have $\abs E=4096$. According to
equation~\eqref{eq:14}, we need to perform the effective measurements
\begin{algorithm}[h]
  \KwData{Pattern $P$, Measurement rate $p$, Image $\bX$, Spectralized
    signature $s$}
  \KwResult{}
  Compute the measurement pattern $E$ that minimizes $\alpha(P)$\;
  Generate a random measurement $f$\;
  Take the effective measurements according to $E+P$\;
  Reconstruct $\bMvirt$ from $\bMeff$\;
  Compute $\bA = \bMvirt^T(\bFvirt\bFvirt^T)^{-1}\bFvirt$\;
  Solve $\displaystyle\argmin_{u\geq 0}\|\phi(u)\|_1
  \quad\text{s.t.}\quad
  \|\bA u-s\|_2<err$\;
  \caption{Compressive Pattern Matching algorithm}
  \label{alg:patterndetect}
\end{algorithm}
denoted by the shifting pattern $E+P$ to be able to reconstruct all
the virtual measurements. The pattern $P$ is rectangular so to
minimize $\alpha(P)$ and according to proposition~\ref{prop:1} the
measurement pattern $E$ should be rectangular-shaped and contained in
a rectangle of height 50 and width 82. That way, we have $\abs E=4096$
and $E+P$, as defined in~\eqref{eq:20}, is obtained by shifting the
pattern $E$ with every element of $P$. The pattern $E+P$ is then
rectangular-shaped and contained in a rectangle of height $50+6-1$ and
width $82+10-1$. More precisely, we have $\abs{E+P}=5001$ which gives
$\alpha(P)\approx1.22$. The effective measurement rate is then
$1.22\times25\approx31\%$ which means that we have to take effective
measurements at a rate of 31\% on the color image to be able to run
the pattern matching minimization problem with a 25\% measurement rate
on the \textit{spectralized} image.

\subsection{Numerical experiments}
\label{sec:numer-exper}

In this section, we illustrate our compressive pattern matching
minimization. The detailed algorithm is described in
Algorithm~\ref{alg:patterndetect}. Given the pattern $P$ and the
measurement rate $p$, we first compute the measurement pattern $E$
that minimizes $\alpha(P)$. For a rectangular-shaped pattern, the
proposition~\ref{prop:1} gives us the optimal solution. For more
complex patterns, if they are compact, they can be approximated by
their enclosing rectangle and the proposition~\ref{prop:1} applies. We
then generate the effective measurements $\bMeff$ from the shifting
pattern $E+P$ and a random measurement $f$. The results of those
measurements are stored in $\bMeff$. We showed that $\bMvirt$ can be
reconstructed from $\bMeff$: it consists essentially in duplicating
and reordering entries of $\bMeff$. However, that process can be
tricky to perform in an efficient way especially in Matlab. For
convenience and testing purposes, we rather build the spectralized
image $\spec_P(\bX)$ and compute $\bMvirt=\bFvirt\spec_P(\bX)$ to more
easily get the matrix $\bMvirt$. Of course, given the size of
$\spec_P(\bX)$ which can be huge, this method shows its limits and we
therefore limit ourselves to images of maximum size $64\times 64$.
The reconstructed $\bMvirt$ allows us to compute the matrix $\bA$ that
is used in our algorithm. To improve the readability of the results,
the images are inverted before display. The algorithm runs in less
than a minute on a classic computer.

\begin{figure}[ht]
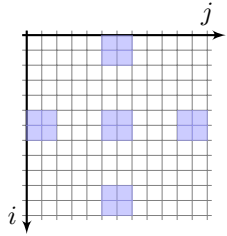

  \centering
  \tikzset{every picture/.style={scale=.2}}%
  \drawpattern{gen/pattern_giza/pattern_giza.raw}
  \caption{Pattern to detect on the Giza image}
  \label{fig:pattern-detect-giza}
\end{figure}

\begin{figure}[ht]
  \centering
  \input{gen/pattern_giza_alpha/eff_wrt_virt.tex}
  \caption{Effective measurement rate with respect to virtual
    measurement rate for the pattern displayed in Fig.~\ref{fig:pattern-detect-giza}}
  \label{fig:effect-meas-rate}
\end{figure}
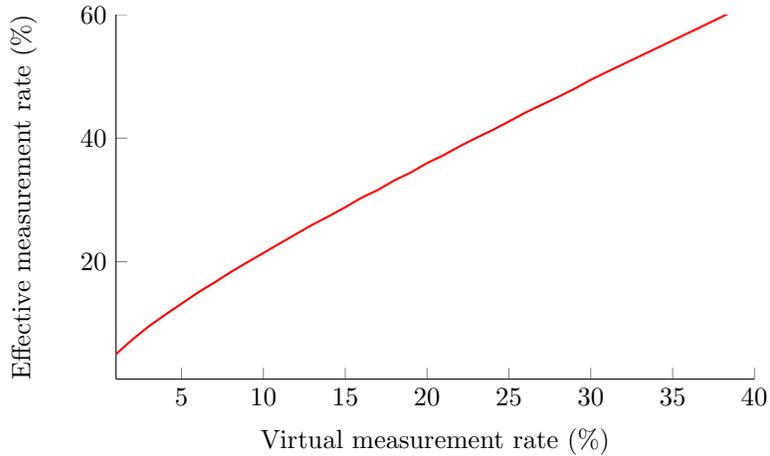

\begin{figure}[ht]
  \centering
  \def\scale{.5}
  \begin{subfigure}[t]{.4\linewidth}
    \centering
      \includegraphics[width=\linewidth]{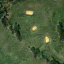}
    \caption{Giza image}
    \label{Giza.Bunker.Pattern.orig.png}
  \end{subfigure}\quad
  \begin{subfigure}[t]{.4\linewidth}
    \centering
      \setlength{\fboxsep}{0pt}%
      \fbox{\includegraphics[width=\linewidth]{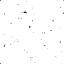}}
    \caption{Compressive pattern matching, 10\% of data}
    \label{Giza.Bunker.Pattern.shift.p10.png.negate.png}
  \end{subfigure}\\
  \begin{subfigure}[t]{.4\linewidth}
    \centering
    \setlength{\fboxsep}{0pt}%
    \fbox{\includegraphics[width=\linewidth]{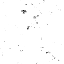}}
    \caption{Compressive pattern matching, 20\% of data}
    \label{Giza.Bunker.Pattern.shift.p20.png.negate.png}
  \end{subfigure}\quad
  \begin{subfigure}[t]{.4\linewidth}
    \centering
    \setlength{\fboxsep}{0pt}%
    \fbox{\includegraphics[width=\linewidth]{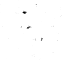}}
    \caption{Compressive pattern matching, 30\% of data}
    \label{Giza.Bunker.Pattern.shift.p30.png.negate.png}
  \end{subfigure}
  \caption{Pattern detection on Peppers image}
  \label{fig:peppers.pattern}
\end{figure}

We first test our algorithm on a publicly available
color image of Giza, Egypt. We extract a $64\times64$ image shown
in Fig.~\ref{Giza.Bunker.Pattern.orig.png} and we want to detect the
locations where there is sand surrounded by vegetation. For that
purpose, we use a pattern whose shape is described in
Fig.~\ref{fig:pattern-detect-giza}.
The centre of the pattern detects the sand and the four other squares
detects the vegetation. The pattern is included in a square of size
$12\times 12$ so we use the measurement pattern as stated in the
proposition~\ref{prop:1}. The minimization problem is first tested for
several virtual measurement rates ranging from 1 to 40 percent.
Figure~\ref{fig:effect-meas-rate} depicts the effective measurement
rate with respect to the virtual measurement rate for the pattern. As
pointed out by proposition~\ref{prop:x-2}, $\alpha$ which is the ratio
of these two values decreases to 1 as the virtual measurements
increases.

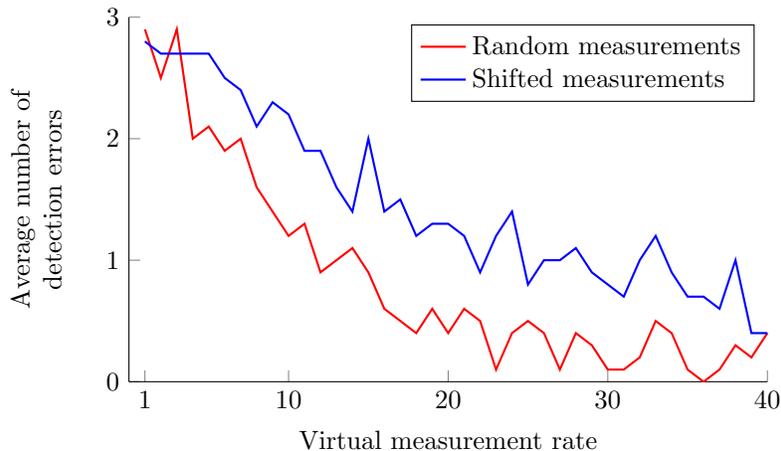
\begin{figure}[ht]
  \centering
  \input{gen/pattern_giza/pattern_errors.tex}
  \caption{Average number of pattern detection errors with respect to
    virtual measurement rate}
  \label{fig:aver-numb-patt}
\end{figure}

Figures~\ref{Giza.Bunker.Pattern.shift.p10.png.negate.png},~\ref{Giza.Bunker.Pattern.shift.p20.png.negate.png}
and~\ref{Giza.Bunker.Pattern.shift.p30.png.negate.png} show the
results of the compressive pattern detection algorithm for a virtual
measurement rate of respectively, 10\%, 20\% and 30\%. The locations
are well recovered for a virtual measurement rate of 20\% and 30\%.
Only one target seems to be detected when a 10\% measurement rate is
taken. More precisely, the graph in Fig.~\ref{fig:aver-numb-patt}
describes the number of pattern detection for various virtual
measurement rate. We test two types of measurements and for each
effective measurement rate we take the average number of errors of 10
minimizations. The random measurements serve as a reference since
using them would require a huge effective measurement rate and thus
totally defeats the purpose of compressive sensing.

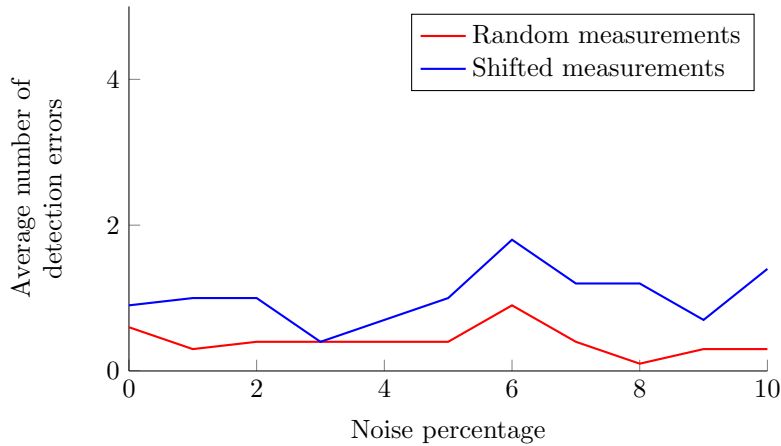
\begin{figure}[ht]
  \centering
  \input{gen/pattern_giza_noise/errors_wrt_noise.tex}
  \caption{Average error detection with respect to noise percentage
    for a fixed measurement rate of 30\%}
  \label{fig:aver-error-detect}
\end{figure}

We then test the same image contaminated with Gaussian noise.
Figure~\ref{fig:aver-error-detect} sums up the our results. We fixed
the virtual measurement rate to 20\% and add Gaussian noise ranging
from 0 to 10\%. The minimization problem appears to be stable when it
comes to Gaussian noise. This is not surprising since the minimization
without noise is operating on $\bM^T(\bF\bF^T)^{-1}\bF$ which is
already a noisy version of the original image $\bX^T$.

On the next numerical experiments, we use a real-world multispectral
image collected by ACTIMAR as part of the HYPLITT project, supported
by the DGA (General Directorate for Armament), France.
See~\cite{jay2012underwater}. The study site is located in Quiberon
Peninsula, on the West coast of France.

\begin{figure}[ht]
  \centering
  \def\scale{.5}
  \begin{subfigure}[t]{.4\linewidth}
    \centering
    \includegraphics[width=\linewidth]{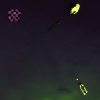}
    \caption{Original image in false color}
    \label{Bache.Carreau.Pattern.orig}
  \end{subfigure}\quad
  \begin{subfigure}[t]{.4\linewidth}
    \centering
    \setlength{\fboxsep}{0pt}%
    \fbox{\includegraphics[width=\linewidth]{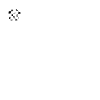}}
    \caption{Non-compressive template detection}
    \label{Bache.Carreau.Pattern.XL1.p30}
  \end{subfigure}\\
  \begin{subfigure}[t]{.4\linewidth}
    \centering
    \setlength{\fboxsep}{0pt}%
    \fbox{\includegraphics[width=\linewidth]{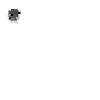}}
    \caption{Compressive template detection, 30\% of data}
    \label{Bache.Carreau.Pattern.TVL1.p30}
  \end{subfigure}
  \quad
  \begin{subfigure}[t]{.4\linewidth}
    \centering
    \setlength{\fboxsep}{0pt}%
    \fbox{\includegraphics[width=\linewidth]{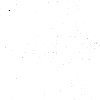}}
    \caption{Compressive pattern detection, 30\% of data}
    \label{Bache.Carreau.Pattern.YALL.p30}
  \end{subfigure}
  \caption{Pattern detection on multispectral image}
  \label{fig:Bache.Carreau.Pattern}
\end{figure}

From the original image of size $316\times 302$ with 160 bands, we
extracted a $100\times 100$ image and selected 12 bands. We wish to
detect the checkered pattern in the top left corner. This time, the
structure of the pattern used is
\ifCLASSOPTIONonecolumn
\begin{equation}
  P=((0, 0), (0, 3), (0, 6), (3, 0), (3, 3), (3, 6), (6, 0), (6, 3), (6, 6))
\end{equation}
\else
\begin{equation}
  \begin{aligned}
    P&=((0, 0), (0, 3), (0, 6),\\
    &\phantom{{}=(}(3, 0), (3, 3), (3, 6),\\
    &\phantom{{}=(}(6, 0), (6, 3), (6, 6))
  \end{aligned}\label{eq:18}
\end{equation}
\fi %
Since the pattern structure is a scaled square, to minimize
$\alpha(P)$ the measurement pattern should be a scaled square as well.
The results of the different minimizations are shown in
Fig.~\ref{fig:Bache.Carreau.Pattern}.
Figure~\ref{Bache.Carreau.Pattern.orig} is the multispectral image in
false color. We first apply the template detection
minimization~\eqref{eq:1} that is able to recover the checkered
structure. The compressive template detection minimization problem in
Fig.~\ref{Bache.Carreau.Pattern.TVL1.p30} fails to detect the
checkered structure. By contrast, the compressive pattern detection
minimization problem for a 30\% measurement rate shown in
Fig.~\ref{Bache.Carreau.Pattern.YALL.p30} clearly detects a checkered
structured as described by the pattern $P$.

\section{Conclusions and perspectives}
\label{sec:conclusion}

After a brief introduction of a new minimization problem from Guo and
Osher that performs template matching on a multispectral image, we
first provide evidence that explains why that minimization problem is
working. Then, we extend it in several ways. We first show that this
minimization problem can be adapted to work with compressive sensing
data. Basically, we obtained comparable results but with limited
number of measurements on the image. Then, we extend the minimization
problem to perform pattern detection with compressive sensing data.
For this purpose, we introduce a formal transformation called the
\textit{spectralization} that depends on the structure of the pattern
we want to detect. Numerical experiments are conducted on both
synthetic and real-world images that validates both approaches. Future
work could consider choosing a sensing matrix that could ease the
computation of the matrix $\bA$ in algorithm~\ref{alg:patterndetect}.
This would allow us to test our algorithms on larger images with a
large number of bands. In particular, we could consider circulant
matrices and orthogonal circulant matrices. One can also study if the
recovery is better if the sensing matrix is built with several
different measurements that are then shifted instead of only one.

\section*{Acknowledgement}
\label{sec:acknowledgement}

This work was supported by the french General Directorate for Armament
(DGA). Moreover, we would like to thank ACTIMAR, which conducted the
field measurement campaign HYPLITT.

We would also like to thank the reviewers for their helpful comments
and insights.

\appendix
\section{Proof of proposition~\ref{prop:1}}
\label{sec:proof-prop-refpr}

\begin{proof}
  Without loss of generality, we suppose
  that the pattern $P$ is defined by
  \begin{equation*}
P=\left\{(i, j),\, 0\leq i<a,\, 0\leq j<b\right\}.
  \end{equation*}
  Let $E$ be a measurement pattern such that $\abs{E}=A$ and let $n_i$
  and $m_i$ be respectively the number of elements of $E+P$ and $E$ of
  ordinate $i$. We have
  \begin{align}
    \label{eq:6}
    \sum_{i\in\mathbb Z}n_i&=\abs{E+P},\\
    \intertext{and}
    \sum_{i\in\mathbb Z}m_i&=\abs{E}.
  \end{align}
  We also remark that for all $i\in\mathbb Z$ and
  $j\in\{i,i-1,\dots,i-a+1\}$, we have
  \begin{equation*}
    \begin{cases}
      m_j=0& \text{if $n_i=0$},\\
      n_i\geq m_j+b-1& \text{if $n_i\neq0$}.
    \end{cases}
  \end{equation*}
  If we set
  \begin{equation}\label{eq:11}
    (\phi_a(u))_i:=\max_{k\in\{i,i-1,\dots,i-a+1\}} u_k,
  \end{equation}
  we can write
  \begin{equation*}
    \begin{cases}
      (\phi_a(m))_i=0& \text{if $n_i=0$},\\
      n_i\geq(\phi_a(m))_i+b-1& \text{if $n_i\neq0$}.
    \end{cases}
  \end{equation*}
  Summing $n_i$ for $i\in\mathbb Z$, we have
  \begin{align}
    \label{eq:7}
    \abs{E+P}&\geq\sum_{i\in\mathbb Z}(\phi_a(m))_i +
    \sum_{\substack{i\in\mathbb Z\\n_i\neq 0}}b-1\\
    &\geq\sum_{i\in\mathbb Z}(\phi_a(m))_i + (\abs{m}+a-1)(b-1).
  \end{align}
  If we take the lower bound on the left side of the inequality for
  all possible measurement patterns $E$ such that $\abs{E}=A$, we have
  \begin{equation}
    \label{eq:8}
    \abs{E+P}\geq\inf_{h\geq1}\left[\inf_{\abs m=h}\sum_{i\in\mathbb Z}(\phi_a(m))_i+ (h+a-1)(b-1)\right].
  \end{equation}
  Using lemma~\ref{lem:1} and plugging it back in~\eqref{eq:8} we have
  \begin{equation}
    \label{eq:10}
    \abs{E+P}\geq A+(a-1)(b-1)+\inf_{h\geq1}(a-1)\ceil{\frac A h}+(b-1)h.
  \end{equation}
\end{proof}

\begin{lemma}\label{lem:1}
  Let $a$ be a non-negative integer, $u$ a non-negative integer valued
  sequence indexed by $\mathbb Z$ such that $\sum_{i\in\mathbb Z}u_i =
  A\geq h$ and $\phi$ defined by~\eqref{eq:11}. We have
  \begin{equation}
    \inf_{\abs u=h}\sum_{i\in\mathbb Z}(\phi_a(u))_i=A+(a-1)\ceil{\frac A h},
  \end{equation}
  and a minimizing sequence is
  \begin{equation*}
    v_i=
    \begin{cases}
      q+1&\text{if $0\leq i<r$},\\
      q&\text{if $r\leq i< h$},\\
      0&\text{otherwise,}
    \end{cases}
  \end{equation*}
  where $q$ and $r$ are respectively the quotient and the remainder in
  the Euclidean division of $A$ by $h$.
\end{lemma}
\begin{proof}
  It is easy to show that $v$ obeys
  \begin{align*}
    \sum_{i\in\mathbb Z}v_i=A\quad\text{and}\quad
    \sum_{i\in\mathbb Z}(\phi_a(v))_i=A+(a-1)\ceil{\frac A h}.
  \end{align*}
  Then, it suffices to show that,
  \begin{equation*}
    \sum_{i\in\mathbb Z}(\phi_a(u))_i\geq\sum_{i\in\mathbb
      Z}(\phi_a(v))_i,
  \end{equation*}
  for all $u$ such that $\abs u=h$ and $\sum_{i\in\mathbb Z}u_i =
  A\geq h$. Without loss of generality we can suppose that the support
  of $u$ is a contiguous subset of $\mathbb Z$ and that this subset is
  $\llbracket 0, h-1\rrbracket$. We then divide our problem into two
  cases:
  \begin{itemize}
    \item If for all $i, j\in\llbracket 0, h-1\rrbracket,
    \abs{u_i-u_j}\leq 1$, we have
    \begin{equation}
      \sum_{i\in\mathbb Z}(\phi_a(u))_i\geq \sum_{i\in\mathbb Z}(\phi_a(v))_i.
    \end{equation}
    \item If not, let $i_0\in\llbracket 0, h-1\rrbracket$ be an index
    such that $u_{i_0}\geq2+u_i$ for all $i\neq i_0$. We choose
    another index $j_0$ distinct from $i_0$ and we construct a new
    sequence $u'$ corresponding to the measurement pattern $E'$ as
    follows
    \begin{equation}
      u'_i=
      \begin{cases}
        u_i-1&\text{if $i=i_0$},\\
        u_i+1&\text{if $i=j_0$},\\
        u_i&\text{otherwise.}
      \end{cases}
    \end{equation}
    We remark that
    \begin{equation*}
      \sum_{i\in\mathbb Z}(\phi_a(u))_i\geq\sum_{i\in\mathbb Z}(\phi_a(u'))_i.
    \end{equation*}
    By induction on $u_{i_0}$ we are reduced to the first case.
  \end{itemize}
  Thus, for all $u$ such that $\abs{u}=h$, we have
  \begin{equation*}
    \sum_{i\in\mathbb Z}(\phi_a(u))_i\geq A+(a-1)\ceil{\frac A h},
  \end{equation*}
  which concludes the proof.
\end{proof}

\bibliographystyle{IEEEtran}
\bibliography{refs.strip.bib}

\end{document}

%% file: preamble-standalone.tex
\usepackage{standalone}
\usepackage[cmex10]{amsmath}
\usepackage{amssymb}
\usepackage{amsthm}

\newtheorem{prop}{Proposition}[section]
\newtheorem{lemma}{Lemma}[section]

\theoremstyle{definition}
\newtheorem{defn}{Definition}

\usepackage[lined, boxed]{algorithm2e}

\usepackage[subnum]{cases}

\usepackage{mathtools}

\usepackage{pgfplotstable}

\usepackage{makecell}

\usepackage{graphicx}
\usepackage{subcaption}

\usepackage{floatrow}

\usepackage{stmaryrd}

\usepackage{grffile}

\usepackage{url}
\usepackage{tikz}
\usetikzlibrary{arrows}
\usetikzlibrary{backgrounds}
\usetikzlibrary{patterns}
\usetikzlibrary{matrix}
\usetikzlibrary{shapes}
\usetikzlibrary{fit}
\usetikzlibrary{calc}
\usetikzlibrary{shadows}
\usetikzlibrary{plotmarks}
\usetikzlibrary{3d}

\usepackage{pgfplots}
\pgfplotsset{%
  every axis plot post/.append style={thick}}

\makeatletter
\tikzoption{canvas is xy plane at z}[]{%
  \def\tikz@plane@origin{\pgfpointxyz{0}{0}{#1}}%
  \def\tikz@plane@x{\pgfpointxyz{1}{0}{#1}}%
  \def\tikz@plane@y{\pgfpointxyz{0}{1}{#1}}%
  \tikz@canvas@is@plane
}
\makeatother

\DeclareMathOperator*{\argmin}{arg\,min}

\DeclareMathOperator*{\sgn}{sgn}

\DeclareMathOperator*{\vecc}{vec}
\DeclareMathOperator{\spec}{spec}
\DeclareMathOperator{\id}{id}
\DeclareMathOperator{\Id}{{\bf Id}}
\DeclareMathOperator{\shrink}{shrink}

\providecommand{\abs}[1]{\left\lvert#1\right\rvert}
\providecommand{\norm}[1]{\left\lVert#1\right\rVert}
\providecommand{\ceil}[1]{\left\lceil#1\right\rceil}

\newcommand{\bphi}{\boldsymbol\phi}
\newcommand{\bPsi}{\boldsymbol\Psi}
\newcommand{\bPhi}{\boldsymbol\Phi}
\newcommand{\bX}{\boldsymbol X}
\newcommand{\bA}{\boldsymbol A}
\newcommand{\bD}{\boldsymbol D}
\newcommand{\bF}{\boldsymbol F}
\newcommand{\bM}{\boldsymbol M}
\newcommand{\bI}{\boldsymbol I}
\newcommand{\bQ}{\boldsymbol Q}
\newcommand{\bB}{\boldsymbol B}
\newcommand{\bMvirt}{\boldsymbol M_{\textbf{virt}}}
\newcommand{\bFvirt}{\boldsymbol F_{\textbf{virt}}}
\newcommand{\bMeff}{\boldsymbol M_{\textbf{eff}}}
\newcommand{\bFeff}{\boldsymbol F_{\textbf{eff}}}

%% file: preamble-local.tex
\input{tikz/tikz-m4.tex}

\def\flag{}

\ifx\flag\undefined
\def\pdflatexcmd{pdflatex --shell-escape --jobname="\jobname\suffix"
  "\string\gdef\string\flag{}\string\input\space\jobname\suffix";
}
\def\cleanfiles{%
  rm "\jobname\suffix.tex"
  rm "\jobname\suffix.bbl"
  rm "\jobname\suffix.blg"
  rm "\jobname\suffix.aux"
  rm "\jobname\suffix.log"
}
\def\suffix{-draft-onecolumn}
\immediate\write18{%
  sed "1 s/final/draftcls, onecolumn/g" "\jobname.tex" > "\jobname\suffix.tex";
  \pdflatexcmd
  bibtex "\jobname\suffix";
  \pdflatexcmd
  \pdflatexcmd
  \cleanfiles
}%
\def\suffix{-final-twocolumn}
\immediate\write18{%
  cat "\jobname.tex" > "\jobname\suffix.tex";
  \pdflatexcmd
  bibtex "\jobname\suffix";
  \pdflatexcmd
  \pdflatexcmd
  \cleanfiles
}%
\fi

\graphicspath{{./img/}}

\usepackage{xr}

%% file: tikz/tikz-m4.tex
\makeatletter
\tikzoption{canvas is xy plane at z}[]{%
  \def\tikz@plane@origin{\pgfpointxyz{0}{0}{#1}}%
  \def\tikz@plane@x{\pgfpointxyz{1}{0}{#1}}%
  \def\tikz@plane@y{\pgfpointxyz{0}{1}{#1}}%
  \tikz@canvas@is@plane
}
\makeatother

\def\foreachxyz#1#2{%
  \pgfplotstableread[header=false, col sep=comma]{#1}\datatable
  \pgfplotstableforeachcolumnelement{[index] 0}\of\datatable\as\x{

    \pgfplotstablegetelem{\pgfplotstablerow}{[index] 1}\of\datatable
    \let\y\pgfplotsretval
    \pgfplotstablegetelem{\pgfplotstablerow}{[index] 2}\of\datatable
    \let\z\pgfplotsretval
    #2}}

\def\foreachxy#1#2{%
  \pgfplotstableread[header=false, col sep=comma]{#1}\datatable
  \pgfplotstableforeachcolumnelement{[index] 0}\of\datatable\as\x{

    \pgfplotstablegetelem{\pgfplotstablerow}{[index] 1}\of\datatable
    \let\y\pgfplotsretval
    #2}}

\tikzset{
  baselinex/.style = { baseline={ ($ #1 + (0, -.5ex) $) } },
  baselinex/.default = (current bounding box.center)
}

\tikzset{
  arrow/.style={->, >=stealth, thick}
}

\tikzset{%
  block/.style = {
    draw,
    rectangle,
    minimum height=3em,
    minimum width=5em,
    rounded corners,
    text centered,
    text opacity=1,
    shade,
    ball color=blue!40!white,
    opacity=0.50},
  warn/.style = {
    block,
    fill=red!20,
    ball color=red!40}}

\newcommand\drawpattern[1]{%
  \begin{tikzpicture}[
    baselinex=(depth),
    arrow/.style = {->, >=latex', thick, shorten >=-7pt}]

    \foreachxy{#1} {%
      \path (\x, \y) rectangle ++(1, 1);
    }

    \coordinate (depth) at (current bounding box.center);

    \coordinate (O) at (0, 0);
    \coordinate (A) at (current bounding box.north east);

    \path (current bounding box.south west) +(-.3, -.3) node (OO) {};
    \path (current bounding box.north east) +(.3, .3) node (AA) {};
    \draw[very thin, gray] (OO) grid (AA);

    \draw[arrow] (O |- AA) -- (O) node[left] {$i$};
    \draw[arrow] (OO |- A) -- (A) node[above] {$j$};

    \foreachxy{#1}{%
      \fill[color=blue!40!white, opacity=.5]
      (\x, \y) rectangle ++(1, 1);
    }
  \end{tikzpicture}
}

\newcommand\pattern[1]{%
  \begin{tikzpicture}[
    baselinex=(depth),
    arrow/.style = {->, >=latex', thick, shorten >=-7pt}]

    \foreach \i in {#1} {
      \path[rotate=-90] \i rectangle ++(1, 1);
    }

    \coordinate (depth) at (current bounding box.center);

    \coordinate (O) at (0, 0);
    \coordinate (A) at (current bounding box.north east);

    \path (current bounding box.south west) +(-.3, -.3) node (OO) {};
    \path (current bounding box.north east) +(.3, .3) node (AA) {};
    \draw[very thin, gray] (OO) grid (AA);

    \draw[arrow] (O |- AA) -- (O |- OO) node[left] {$i$};
    \draw[arrow] (OO |- A) -- (AA |- O) node[above] {$j$};

    \foreach \i in {#1} {
      \fill[color=blue!40!white, opacity=.5, rotate=-90]
      \i rectangle ++(1, 1);
    }

  \end{tikzpicture}
}

%% file: gen/moffet_template.tex
\newcommand\moffeterror{%
  \hline
  10&0.37&3.62\\
  \hline
  20&0.67&2.50\\
  \hline
  30&1.16&1.64\\
  \hline
  40&1.40&1.53\\
  \hline
}

\newcommand\moffetXLun{%
  \hline
  1.56&0.00\\
  \hline
}

%% file: tikz/NormeL1_CS.tex
\newcommand\droite[2]{%
  \draw ($ ($ #1 $)!20!($ #2 $) $) -- ($ ($ #1 $)!-20!($ #2 $) $)}
\begin{tikzpicture}

  \coordinate (A) at (-1.2, -1.2);
  \coordinate (B) at (1.5, 1.5);
  \coordinate (O) at (0, 0);

  \draw[->, >=latex'] (A |- O)  -- (B |- O) node[below] {$u_1$};
  \draw[->, >=latex'] (A -| O)  -- (B -| O) node[right] {$u_2$};

  \fill[blue!50!white, fill=blue!40!white, opacity=.6]
  (1, 0) -- (0, 1) -- (-1, 0) -- (0, -1) -- cycle;

  \clip (A) rectangle (B);

  \droite{(0, 1)}{(4, 0)};

  \fill[red, thick] (0, 1) circle[radius=1pt];
\end{tikzpicture}
\ifstandalone

\fi

%% file: tikz/NormeL1_Template.tex
\newcommand\droite[2]{%
  \draw ($ ($ #1 $)!20!($ #2 $) $) -- ($ ($ #1 $)!-20!($ #2 $) $)}
\begin{tikzpicture}

  \coordinate (A) at (-1.2, -1.2);
  \coordinate (B) at (1.5, 1.5);
  \coordinate (O) at (0, 0);

  \draw[->, >=latex'] (A |- O)  -- (B |- O) node[below] {$u_1$};
  \draw[->, >=latex'] (A -| O)  -- (B -| O) node[right] {$u_2$};

  \fill[blue!50!white, fill=blue!40!white, opacity=.6]
  (1, 0) -- (0, 1) -- (-1, 0) -- (0, -1) -- cycle;

  \clip (A) rectangle (B);

  \droite{(0, 1)}{(1, 0)};

  \fill[red, thick] (0, 1) circle[radius=1pt];
  \fill[red, thick] (1, 0) circle[radius=1pt];
  \draw[red, thick] (0, 1) -- (1, 0);
\end{tikzpicture}
\ifstandalone

\fi

%% file: gen/genlambda.tex
%
%
%
%
\begin{tikzpicture}

\begin{axis}[%
view={0}{90},
width=0.6\linewidth,
height=0.4248563049853372\linewidth,
scale only axis,
xmin=1, xmax=5,
xlabel={$\lambda$},
ymin=0.5, ymax=1.4,
ylabel={$\|\lambda F^T(FF^T)^{-1}F-I_{n_p}\|_\infty$}]
\addplot [
color=red,
solid,
forget plot
]
coordinates{
 (1,0.841812896943765)(1.04,0.834817277862225)(1.08,0.832539399708194)(1.12,0.826886742328599)(1.16,0.820304299475323)(1.2,0.813086117620093)(1.24,0.811896723594062)(1.28,0.797327590412994)(1.32,0.79561641735747)(1.36,0.784603149574253)(1.4,0.780657247835821)(1.44,0.777662247826312)(1.48,0.767843022012177)(1.52,0.761424005996615)(1.56,0.759397698993218)(1.6,0.746341857874074)(1.64,0.743698402648094)(1.68,0.738283153963682)(1.72,0.729168100755469)(1.76,0.726156432743921)(1.8,0.719216570470034)(1.84,0.712358455941968)(1.88,0.708813927704489)(1.92,0.706813240008831)(1.96,0.69481823726303)(2,0.693051407677543)(2.04,0.678124504794353)(2.08,0.682581167273599)(2.12,0.678401510706784)(2.16,0.658997076322919)(2.2,0.653912387398265)(2.24,0.656861319061199)(2.28,0.635584166904158)(2.32,0.624805185809672)(2.36,0.633115551270531)(2.4,0.626628548034333)(2.44,0.623688946223259)(2.48,0.617683014533301)(2.52,0.613867676513465)(2.56,0.601599458898178)(2.6,0.586275802764052)(2.64,0.581345395413699)(2.68,0.575896013102724)(2.72,0.579324010913194)(2.76,0.566312862933145)(2.8,0.56774271432453)(2.84,0.563844077058398)(2.88,0.55991006850333)(2.92,0.558862934170319)(2.96,0.551957080466053)(3,0.555175038005845)(3.04,0.565400466685253)(3.08,0.567397994225414)(3.12,0.568086914883156)(3.16,0.579664927010107)(3.2,0.582416702345751)(3.24,0.603075350955548)(3.28,0.614823404468434)(3.32,0.609526496520322)(3.36,0.628968001708064)(3.4,0.645666406284367)(3.44,0.65045277958286)(3.48,0.666828719477801)(3.52,0.69354852046481)(3.56,0.70392848153797)(3.6,0.720581089271258)(3.64,0.734457612027198)(3.68,0.758444751971315)(3.72,0.765694788846911)(3.76,0.767855970684693)(3.8,0.798729798119124)(3.84,0.811904257545716)(3.88,0.850093351699045)(3.92,0.865290890374334)(3.96,0.86679956027676)(4,0.900862026727311)(4.04,0.910415537067262)(4.08,0.933382579191547)(4.12,0.967375585204145)(4.16,0.982991613795595)(4.2,0.96538836809324)(4.24,0.994854841085596)(4.28,1.02484460546959)(4.32,1.06384756301626)(4.36,1.05949524581873)(4.4,1.07845028005617)(4.44,1.07787710544699)(4.48,1.12068702790204)(4.52,1.1643355550526)(4.56,1.15764919175492)(4.6,1.20132226268323)(4.64,1.20668923424932)(4.68,1.22768777026156)(4.72,1.23020851381608)(4.76,1.26013287832155)(4.8,1.26990415915857)(4.84,1.28118770064076)(4.88,1.31294832652893)(4.92,1.32404611706904)(4.96,1.34718910795463)(5,1.3766028054091)
};
\end{axis}
\end{tikzpicture}%

%% file: gen/generrmax.tex
%
%
%
%

\definecolor{mycolor1}{rgb}{1,0,1}

\begin{tikzpicture}

\begin{axis}[%
view={0}{90},
width=0.8\linewidth,
height=0.66417399804497\linewidth,
scale only axis,
xmin=0.1, xmax=0.5,
xlabel={$p=\frac m {n_P}$},
ymin=0.2, ymax=1.6,
ylabel={$\|\cdot\|_\infty$},
legend style={align=left}]
\addplot [
color=red,
solid
]
coordinates{
 (0.1,1.56997486826808)(0.11,1.45304441503425)(0.12,1.4188841393839)(0.13,1.34980010015798)(0.14,1.31635502931885)(0.15,1.23303024585903)(0.16,1.20401338785258)(0.17,1.18356500097182)(0.18,1.13798144502851)(0.19,1.05982210180146)(0.2,1.049750891708)(0.21,1.01227394262077)(0.22,0.989198800902672)(0.23,0.939096937725131)(0.24,0.924769598934187)(0.25,0.896934095235396)(0.26,0.888332729961813)(0.27,0.892055049616554)(0.28,0.86485028809249)(0.29,0.864026548993869)(0.3,0.839329850624864)(0.31,0.813913910746742)(0.32,0.796632257830884)(0.33,0.781092584411892)(0.34,0.778702452582932)(0.35,0.762822397693593)(0.36,0.774242793531457)(0.37,0.746286639519722)(0.38,0.718364951298617)(0.39,0.711625204893827)(0.4,0.700755710851934)(0.41,0.709333521969348)(0.42,0.685456609072853)(0.43,0.674386676646437)(0.44,0.665350468324655)(0.45,0.668521181224541)(0.46,0.645695287234458)(0.47,0.650752499264049)(0.48,0.641193756987924)(0.49,0.635654449430828)(0.5,0.613411705377322)
};
\addlegendentry{T1, iid Gaussian};

\addplot [
color=blue,
solid
]
coordinates{
 (0.1,1.39656919375115)(0.11,1.28294214986302)(0.12,1.24433891665175)(0.13,1.18264785331212)(0.14,1.09566285912226)(0.15,1.0461609396164)(0.16,1.01781282495241)(0.17,0.980616810719017)(0.18,0.923045080832468)(0.19,0.908098272697357)(0.2,0.876866981898231)(0.21,0.816974799630216)(0.22,0.796087792610521)(0.23,0.758264632394939)(0.24,0.727959476867649)(0.25,0.71055502273964)(0.26,0.706205104532034)(0.27,0.67547139928555)(0.28,0.651530697604203)(0.29,0.666848761970322)(0.3,0.628406908211096)(0.31,0.613568659854237)(0.32,0.605152745368751)(0.33,0.570236361256097)(0.34,0.565243224867147)(0.35,0.554388495692015)(0.36,0.529812654927931)(0.37,0.516513670966016)(0.38,0.509176863028649)(0.39,0.494234562828514)(0.4,0.493203318203096)(0.41,0.484785316419395)(0.42,0.466123026069271)(0.43,0.45521731894434)(0.44,0.447689720577431)(0.45,0.433936865701554)(0.46,0.430101408333758)(0.47,0.421489868163376)(0.48,0.409084552868374)(0.49,0.400690621649935)(0.5,0.397828167795676)
};
\addlegendentry{T2, iid Gaussian};

\addplot [
color=mycolor1,
solid
]
coordinates{
 (0.1,1.31741007516348)(0.11,1.30891886509771)(0.12,1.15397822108937)(0.13,1.12780138608398)(0.14,1.03995298988945)(0.15,0.984435323893216)(0.16,0.955776978043337)(0.17,0.935725978957276)(0.18,0.932337782017874)(0.19,0.851442945093841)(0.2,0.838926570497285)(0.21,0.805421048194578)(0.22,0.802599838417668)(0.23,0.769175864197459)(0.24,0.788311892247756)(0.25,0.712499183121576)(0.26,0.702096892957193)(0.27,0.694411200242406)(0.28,0.676366349879397)(0.29,0.656551214411033)(0.3,0.653960184548453)(0.31,0.640559793084155)(0.32,0.636220178629464)(0.33,0.633752348226828)(0.34,0.595227894033824)(0.35,0.597107075557418)(0.36,0.602101425221348)(0.37,0.556459728182664)(0.38,0.574725762141995)(0.39,0.560730434830436)(0.4,0.537045072142122)(0.41,0.546768948445301)(0.42,0.535936636413587)(0.43,0.524030844269794)(0.44,0.517933265352823)(0.45,0.502504359694448)(0.46,0.508474355991824)(0.47,0.495885028553785)(0.48,0.501831654537442)(0.49,0.48662938165892)(0.5,0.479877883267275)
};
\addlegendentry{T1, iid Gaussian circulant};

\addplot [
color=black,
solid
]
coordinates{
 (0.1,1.22614296683138)(0.11,1.15291247236512)(0.12,1.05194889161414)(0.13,1.02084172333999)(0.14,0.95089629052918)(0.15,0.910054868227374)(0.16,0.874280362389947)(0.17,0.844892295357799)(0.18,0.811699007949385)(0.19,0.761988136247715)(0.2,0.742261876840081)(0.21,0.706171486026747)(0.22,0.70818227466381)(0.23,0.672752666706702)(0.24,0.653882422666218)(0.25,0.6271392340168)(0.26,0.610668552546861)(0.27,0.589688310162752)(0.28,0.570441938964987)(0.29,0.564549735604504)(0.3,0.542341721060119)(0.31,0.533625089504712)(0.32,0.519989703789067)(0.33,0.509215591428367)(0.34,0.487437134548619)(0.35,0.491633907671329)(0.36,0.480444326785001)(0.37,0.47137190210081)(0.38,0.460135767692495)(0.39,0.449216694879536)(0.4,0.435304173632808)(0.41,0.425817636163823)(0.42,0.419810097232355)(0.43,0.408320205559708)(0.44,0.400397275803957)(0.45,0.386164076231593)(0.46,0.384307609658411)(0.47,0.374832535839517)(0.48,0.367117068092849)(0.49,0.367509447207359)(0.5,0.353886935714753)
};
\addlegendentry{T2, iid Gaussian circulant};

\end{axis}
\end{tikzpicture}%

%% file: gen/template_giza/template_giza_plot.tex
%
%
%

\definecolor{mycolor1}{rgb}{1,0,1}%

\begin{tikzpicture}

\begin{axis}[%
  width=0.7\textwidth,
height=0.6\textwidth,
scale only axis,
xmin=0,
xmax=40,
xlabel={measurement rate (\%)},
ymin=0,
ymax=3.5,
ylabel={wrong detection (\%)},
axis x line*=bottom,
axis y line*=left,
legend columns=2,
legend style={draw=black,fill=white,legend cell align=left, anchor=center,
  at={(.5, -.35)}}
]
\addplot [
color=black,
solid
]
table[row sep=newline]{gen/template_giza/L1_circ.table};
\addlegendentry{$L_1$, circulant};

\addplot [
color=green,
solid
]
table[row sep=newline]{gen/template_giza/TVL1_circ.table};
\addlegendentry{$TV/L_1$, circulant};

\addplot [
color=red,
solid
]
table[row sep=newline]{gen/template_giza/L1.table};
\addlegendentry{$L_1$, Gaussian};

\addplot [
color=blue,
solid
]
table[row sep=newline]{gen/template_giza/TVL1.table};
\addlegendentry{$TV/L_1$, Gaussian};

\addplot [
color=mycolor1,
solid
]
table[row sep=crcr]{
1 0.03\\
2 0.03\\
3 0.03\\
4 0.03\\
5 0.03\\
6 0.03\\
7 0.03\\
8 0.03\\
9 0.03\\
10 0.03\\
11 0.03\\
12 0.03\\
13 0.03\\
14 0.03\\
15 0.03\\
16 0.03\\
17 0.03\\
18 0.03\\
19 0.03\\
20 0.03\\
21 0.03\\
22 0.03\\
23 0.03\\
24 0.03\\
25 0.03\\
26 0.03\\
27 0.03\\
28 0.03\\
29 0.03\\
30 0.03\\
31 0.03\\
32 0.03\\
33 0.03\\
34 0.03\\
35 0.03\\
36 0.03\\
37 0.03\\
38 0.03\\
39 0.03\\
40 0.03\\
};
\addlegendentry{Template matching};
\end{axis}
\end{tikzpicture}%


%% file: gen/moffett_signature.tex
%
%
%
\begin{tikzpicture}

\begin{axis}[%
width=0.4\linewidth,
height=0.315483870967742\linewidth,
scale only axis,
xmin=0,
xmax=16,
xlabel={Spectral bands},
ymin=0.2,
ymax=0.9,
ylabel={Intensity}
]
\addplot [
color=blue,
solid,
forget plot
]
table[row sep=crcr]{
1 0.612408836275235\\
2 0.630483035619913\\
3 0.791776767783532\\
4 0.788817249762182\\
5 0.781735545925378\\
6 0.782475425430716\\
7 0.767994926540535\\
8 0.763978437797273\\
9 0.70415389493711\\
10 0.70225134763767\\
11 0.696649402811542\\
12 0.695909523306204\\
13 0.558397632385583\\
14 0.556495085086143\\
15 0.268364866293204\\
16 0.267307895571293\\
};
\end{axis}
\end{tikzpicture}%

%% file: gen/signature_moffet/signature_moffett_plot.tex
%
%
%

\definecolor{mycolor1}{rgb}{1,0,1}%

\begin{tikzpicture}

\begin{axis}[%
width=0.6\textwidth,
height=0.6\textwidth,
scale only axis,
xmin=0,
xmax=40,
xlabel={measurement rate (\%)},
ymin=0,
ymax=8,
ylabel={wrong detection (\%)},
axis x line*=bottom,
axis y line*=left,
legend columns=2,
legend style={draw=black,fill=white,legend cell align=left, anchor=center,
  at={(.5, -.35)}}
]
\addplot [
color=black,
solid
]
table[row sep=crcr]{
1 7.71728519\\
2 7.29492188\\
3 6.41601562\\
4 6.0620117\\
5 7.11181641\\
6 6.51855471\\
7 7.10449217\\
8 6.50390624\\
9 5.9399414\\
10 6.04492186\\
11 6.14990234\\
12 5.85449218\\
13 6.31347656\\
14 5.56396485\\
15 5.41015625\\
16 5.49560547\\
17 6.37451173\\
18 5.59326172\\
19 5.84472656\\
20 5.38085937\\
21 5.31982422\\
22 5.70312499\\
23 5.62255859\\
24 5.45654296\\
25 5.63232422\\
26 5.10498046\\
27 5.87402342\\
28 5.41992187\\
29 5.09521485\\
30 5.49560548\\
31 5.64208984\\
32 5.1904297\\
33 5.60791015\\
34 5.08789061\\
35 5.3857422\\
36 5.16357421\\
37 4.96826171\\
38 4.98046874\\
39 4.90722656\\
40 4.72900391\\
};
\addlegendentry{$L_1$, circulant};

\addplot [
color=green,
solid
]
table[row sep=crcr]{
1 7.07031248\\
2 5.85693358\\
3 5.74951173\\
4 4.94384765\\
5 5.47363282\\
6 5.47851561\\
7 6.33300781\\
8 5.53955079\\
9 4.8046875\\
10 5.74951173\\
11 6.15722656\\
12 4.99511719\\
13 5.22216796\\
14 5.37353517\\
15 4.55078125\\
16 5.30029297\\
17 4.60205078\\
18 5.16601561\\
19 5.82763673\\
20 5.42968751\\
21 5.72021484\\
22 4.43847657\\
23 5.17333984\\
24 5.19775393\\
25 4.95849608\\
26 5.50292969\\
27 4.65087891\\
28 4.54101562\\
29 5.29052735\\
30 4.78027343\\
31 4.06982423\\
32 5.2099609\\
33 4.65332032\\
34 4.50683594\\
35 4.29199217\\
36 4.72412109\\
37 5.3149414\\
38 4.74609375\\
39 3.96728516\\
40 4.32128905\\
};
\addlegendentry{$TV/L_1$, circulant};

\addplot [
color=red,
solid
]
table[row sep=crcr]{
1 5.24658201\\
2 5.38330079\\
3 5.51513672\\
4 5.61279298\\
5 5.37841798\\
6 5.22705077\\
7 5.05126955\\
8 5.51269532\\
9 5.07080078\\
10 5.28320313\\
11 5.45410157\\
12 5.20996093\\
13 5.11474608\\
14 5.13916016\\
15 5.33691405\\
16 5.26367187\\
17 4.85107421\\
18 5.16357421\\
19 4.95849608\\
20 5.12207032\\
21 4.88525389\\
22 5\\
23 4.82421873\\
24 5.02929688\\
25 4.9633789\\
26 5.07568358\\
27 5.02441405\\
28 5.01953124\\
29 5.13183594\\
30 5.01220704\\
31 5.20263673\\
32 4.99511717\\
33 5.05859375\\
34 4.99755858\\
35 5.12695312\\
36 5.2758789\\
37 5\\
38 5.18066406\\
39 5.20751952\\
40 5.17089843\\
};
\addlegendentry{$L_1$, Gaussian};

\addplot [
color=blue,
solid
]
table[row sep=crcr]{
1 4.64355468\\
2 4.97314453\\
3 4.52148436\\
4 4.90966797\\
5 4.73632812\\
6 4.79736328\\
7 4.2578125\\
8 5.02197266\\
9 4.4067383\\
10 4.60449219\\
11 4.60937501\\
12 4.29443359\\
13 4.58740235\\
14 5.49804687\\
15 4.81201171\\
16 5.13916016\\
17 4.81933593\\
18 5.15380859\\
19 5.0415039\\
20 4.66552735\\
21 4.08447266\\
22 5.37597656\\
23 4.90966796\\
24 5.52001953\\
25 5.44189453\\
26 5.51513672\\
27 4.95605468\\
28 5.02441408\\
29 4.32617186\\
30 4.74121093\\
31 4.29199219\\
32 4.83886718\\
33 4.7338867\\
34 5.0732422\\
35 5.11230469\\
36 5.3149414\\
37 4.46777343\\
38 5.09521484\\
39 4.63623047\\
40 4.3774414\\
};
\addlegendentry{$TV/L_1$, Gaussian};

\addplot [
color=mycolor1,
solid
]
table[row sep=crcr]{
1 0.3\\
2 0.3\\
3 0.3\\
4 0.3\\
5 0.3\\
6 0.3\\
7 0.3\\
8 0.3\\
9 0.3\\
10 0.3\\
11 0.3\\
12 0.3\\
13 0.3\\
14 0.3\\
15 0.3\\
16 0.3\\
17 0.3\\
18 0.3\\
19 0.3\\
20 0.3\\
21 0.3\\
22 0.3\\
23 0.3\\
24 0.3\\
25 0.3\\
26 0.3\\
27 0.3\\
28 0.3\\
29 0.3\\
30 0.3\\
31 0.3\\
32 0.3\\
33 0.3\\
34 0.3\\
35 0.3\\
36 0.3\\
37 0.3\\
38 0.3\\
39 0.3\\
40 0.3\\
};
\addlegendentry{Template matching};

\end{axis}
\end{tikzpicture}%

%% file: gen/pattern_giza_alpha/eff_wrt_virt.tex
\begin{tikzpicture}
  \begin{axis}[%
    width=.7\linewidth,
    height=0.4\linewidth,
    scale only axis,
    xmin=1,
    xmax=40,
    xlabel={Virtual measurement rate (\%)},
    ymin=1,
    ymax=60,
    ylabel style={align=center},
    ylabel={Effective measurement rate (\%)},
    axis x line*=bottom,
    axis y line*=left,
    legend style={draw=black,fill=white,legend cell align=left}
    ]
    \addplot[color=red, solid]
    table[x index=0,y index=1,header=false]
    {gen/pattern_giza_alpha/data.dat};
  \end{axis}
\end{tikzpicture}


%% file: gen/pattern_giza/pattern_errors.tex
\begin{tikzpicture}
  \begin{axis}[%
    width=.7\linewidth,
    height=0.4\linewidth,
    scale only axis,
    xmin=0,
    xmax=40,
    xtick={1, 10, 20, 30, 40},
    xlabel={Virtual measurement rate},
    ymin=0,
    ymax=3,
    ylabel style={align=center},
    ylabel=Average number of\\detection errors,
    axis x line*=bottom,
    axis y line*=left,
    legend style={draw=black,fill=white,legend cell align=left}
    ]
    \addplot[
    color=red,
    solid]
    table[x index=0,y index=1,header=false]
    {gen/pattern_giza/errors_rand_wrt_meas_rate.raw};
    \addlegendentry{Random measurements};

    \addplot[
    color=blue,
    solid]
    table[x index=0,y index=1,header=false]
    {gen/pattern_giza/errors_shift_wrt_meas_rate.raw};
    \addlegendentry{Shifted measurements};
  \end{axis}
\end{tikzpicture}


%% file: gen/pattern_giza_noise/errors_wrt_noise.tex
\begin{tikzpicture}
  \begin{axis}[%
    width=.7\linewidth,
    height=0.4\linewidth,
    scale only axis,
    xmin=0,
    xmax=10,
    xlabel={Noise percentage},
    ymin=0,
    ymax=5,
    ylabel style={align=center},
    ylabel=Average number of\\detection errors,
    axis x line*=bottom,
    axis y line*=left,
    legend style={draw=black,fill=white,legend cell align=left}
    ]
    \addplot[
    color=red,
    solid]
    table[x index=0,y index=1,header=false]
    {gen/pattern_giza_noise/errors_rand_wrt_noise.raw};
    \addlegendentry{Random measurements};

    \addplot[
    color=blue,
    solid]
    table[x index=0,y index=1,header=false]
    {gen/pattern_giza_noise/errors_shift_wrt_noise.raw};
    \addlegendentry{Shifted measurements};

  \end{axis}
\end{tikzpicture}


%% file: HAL_article.bbl
\begin{thebibliography}{10}
\providecommand{\url}[1]{#1}
\csname url@samestyle\endcsname
\providecommand{\newblock}{\relax}
\providecommand{\bibinfo}[2]{#2}
\providecommand{\BIBentrySTDinterwordspacing}{\spaceskip=0pt\relax}
\providecommand{\BIBentryALTinterwordstretchfactor}{4}
\providecommand{\BIBentryALTinterwordspacing}{\spaceskip=\fontdimen2\font plus
\BIBentryALTinterwordstretchfactor\fontdimen3\font minus
  \fontdimen4\font\relax}
\providecommand{\BIBforeignlanguage}[2]{{%
\expandafter\ifx\csname l@#1\endcsname\relax
\typeout{** WARNING: IEEEtran.bst: No hyphenation pattern has been}%
\typeout{** loaded for the language `#1'. Using the pattern for}%
\typeout{** the default language instead.}%
\else
\language=\csname l@#1\endcsname
\fi
#2}}
\providecommand{\BIBdecl}{\relax}
\BIBdecl

\bibitem{donoho2006compressed}
D.~Donoho, ``Compressed sensing,'' \emph{Information Theory, IEEE Transactions
  on}, vol.~52, no.~4, pp. 1289--1306, 2006.

\bibitem{candes2006stable}
E.~Cand\`es, J.~Romberg, and T.~Tao, ``Stable signal recovery from incomplete
  and inaccurate measurements,'' \emph{Communications on pure and applied
  mathematics}, vol.~59, no.~8, pp. 1207--1223, 2006.

\bibitem{mallat1993matching}
S.~Mallat and Z.~Zhang, ``{Matching pursuits with time-frequency
  dictionaries},'' \emph{IEEE Transactions on signal processing}, vol.~41,
  no.~12, pp. 3397--3415, 1993.

\bibitem{pati1993orthogonal}
Y.~C. Pati, R.~Rezaiifar, and P.~Krishnaprasad, ``Orthogonal matching pursuit:
  Recursive function approximation with applications to wavelet
  decomposition,'' in \emph{Signals, Systems and Computers, 1993. 1993
  Conference Record of The Twenty-Seventh Asilomar Conference on}.\hskip 1em
  plus 0.5em minus 0.4em\relax IEEE, 1993, pp. 40--44.

\bibitem{Tropp07signalrecovery}
J.~A. Tropp, Anna, and C.~Gilbert, ``Signal recovery from random measurements
  via orthogonal matching pursuit,'' \emph{IEEE Trans. Inform. Theory},
  vol.~53, pp. 4655--4666, 2007.

\bibitem{blumensath2008gradient}
T.~Blumensath and M.~E. Davies, ``Gradient pursuits,'' \emph{Signal Processing,
  IEEE Transactions on}, vol.~56, no.~6, pp. 2370--2382, 2008.

\bibitem{davis1997adaptive}
G.~Davis, S.~Mallat, and M.~Avellaneda, ``Adaptive greedy approximations,''
  \emph{Constructive approximation}, vol.~13, no.~1, pp. 57--98, 1997.

\bibitem{donoho2006stable}
D.~L. Donoho, M.~Elad, and V.~N. Temlyakov, ``Stable recovery of sparse
  overcomplete representations in the presence of noise,'' \emph{Information
  Theory, IEEE Transactions on}, vol.~52, no.~1, pp. 6--18, 2006.

\bibitem{Chen98atomicdecomposition}
S.~S. Chen, D.~L. Donoho, and M.~A. Saunders, ``Atomic decomposition by basis
  pursuit,'' \emph{SIAM Journal on Scientific Computing}, vol.~20, pp. 33--61,
  1998.

\bibitem{duarte2008single}
M.~F. Duarte, M.~A. Davenport, D.~Takhar, J.~N. Laska, T.~Sun, K.~F. Kelly, and
  R.~G. Baraniuk, ``Single-pixel imaging via compressive sampling,''
  \emph{Signal Processing Magazine, IEEE}, vol.~25, no.~2, pp. 83--91, 2008.

\bibitem{gehm2007single}
M.~Gehm, R.~John, D.~Brady, R.~Willett, T.~Schulz \emph{et~al.}, ``Single-shot
  compressive spectral imaging with a dual-disperser architecture,'' \emph{Opt.
  Express}, vol.~15, no.~21, pp. 14\,013--14\,027, 2007.

\bibitem{davenport2006detection}
M.~A. Davenport, M.~B. Wakin, and R.~G. Baraniuk, ``Detection and estimation
  with compressive measurements,'' \emph{Dept. of ECE, Rice University, Tech.
  Rep}, 2006.

\bibitem{wang2008subspace}
Z.~Wang, G.~R. Arce, and B.~M. Sadler, ``Subspace compressive detection for
  sparse signals,'' in \emph{Acoustics, Speech and Signal Processing, 2008.
  ICASSP 2008. IEEE International Conference on}.\hskip 1em plus 0.5em minus
  0.4em\relax IEEE, 2008, pp. 3873--3876.

\bibitem{li2011compressive}
C.~Li, T.~Sun, K.~Kelly, and Y.~Zhang, ``A compressive sensing and unmixing
  scheme for hyperspectral data processing,'' \emph{Image Processing, IEEE
  Transactions on}, no.~99, pp. 1--1, 2011.

\bibitem{zare2012directly}
A.~Zare, P.~Gader, and K.~S. Gurumoorthy, ``Directly measuring material
  proportions using hyperspectral compressive sensing,'' \emph{Geoscience and
  Remote Sensing Letters, IEEE}, vol.~9, no.~3, pp. 323--327, 2012.

\bibitem{golbabaee2010multichannel}
M.~Golbabaee, S.~Arberet, P.~Vandergheynst \emph{et~al.}, ``Multichannel
  compressed sensing via source separation for hyperspectral images,'' in
  \emph{Eusipco 2010}, 2010.

\bibitem{rousseau2013compressive}
S.~Rousseau, D.~Helbert, P.~Carr\'e, and J.~Blanc-Talon, ``Compressive template
  matching on multispectral data,'' in \emph{International Conference on
  Acoustics, Speech, and Signal Processing (ICASSP)}.\hskip 1em plus 0.5em
  minus 0.4em\relax Vancouver, Canada: IEEE, 2013.

\bibitem{guo2011template}
Z.~Guo and S.~Osher, ``{Template Matching via L1 Minimization and Its
  Application to Hyperspectral Data},'' \emph{Inverse Problems and Imaging},
  vol.~5, no.~1, pp. 19--35, 2011.

\bibitem{goldstein2009split}
T.~Goldstein and S.~Osher, ``{The Split Bregman Method for L1-Regularized
  Problems},'' \emph{SIAM Journal on Imaging Sciences}, vol.~2, p. 323, 2009.

\bibitem{cai2009split}
J.~Cai, S.~Osher, and Z.~Shen, ``Split bregman methods and frame based image
  restoration,'' \emph{Multiscale Model. Simul}, vol.~8, no.~2, pp. 337--369,
  2009.

\bibitem{wang2007fast}
Y.~Wang and W.~Yin, ``A fast algorithm for image deblurring with total
  variation regularization,'' \emph{Image Rochester NY}, pp. 1--19, 2007.

\bibitem{romberg2009compressive}
J.~Romberg, ``Compressive sensing by random convolution,'' \emph{SIAM Journal
  on Imaging Sciences}, vol.~2, no.~4, pp. 1098--1128, 2009.

\bibitem{rudelson1999random}
M.~Rudelson, ``Random vectors in the isotropic position,'' \emph{Journal of
  Functional Analysis}, vol. 164, no.~1, pp. 60--72, 1999.

\bibitem{yin2010practical}
W.~Yin, S.~Morgan, J.~Yang, and Y.~Zhang, ``Practical compressive sensing with
  toeplitz and circulant matrices,'' \emph{Rice University CAAM Technical
  Report TR10-01}, vol.~1, 2010.

\bibitem{max1960quantizing}
J.~Max, ``Quantizing for minimum distortion,'' \emph{IEEE Transactions on
  Information Theory}, vol.~6, no.~1, pp. 7--12, 1960.

\bibitem{jay2012underwater}
S.~Jay, M.~Guillaume, and J.~Blanc-Talon, ``Underwater target detection with
  hyperspectral data: Solutions for both known and unknown water quality,''
  \emph{Selected Topics in Applied Earth Observations and Remote Sensing, IEEE
  Journal of}, vol.~5, no.~4, pp. 1213--1221, 2012.

\end{thebibliography}
